%% file: main.tex
\documentclass[10pt,twocolumn,letterpaper]{article}

\usepackage[pagenumbers]{cvpr}

\input{preamble}

\definecolor{linkblue}{rgb}{0.21,0.49,0.74}
\usepackage[pagebackref,breaklinks,colorlinks,allcolors=linkblue]{hyperref}

\title{When Does an AUROC Comparison Survive a Change in Population Mix?}

\author{Mahtab Movahhedrad\\
University of Southern California
\and
Hamidreza Hasani Balyani\\
Independent Researcher
\and
Seyed Pouyan Mousavi Davoudi\\
Islamic Azad University, Amol Branch
}

\begin{document}
\maketitle
\begin{NoHyper}\renewcommand{\thefootnote}{}\footnotetext{This work has been submitted to the IEEE for possible publication. Copyright may be transferred without notice, after which this version may no longer be accessible.}\end{NoHyper}
\input{sec/0_abstract}
\input{sec/1_intro}
\input{sec/2_formulation}
\input{sec/3_limits}
\input{sec/4_audit}
\input{sec/5_experiments}
\input{sec/6_conclusion}
\input{sec/7_llm_statement}
{
    \small
    \bibliographystyle{ieeenat_fullname}
    \bibliography{references}
}

\clearpage
\onecolumn
\appendix

\setcounter{theorem}{0}
\setcounter{equation}{0}
\setcounter{table}{0}
\setcounter{figure}{0}
\renewcommand{\thetheorem}{S\arabic{theorem}}
\renewcommand{\theequation}{S\arabic{equation}}
\renewcommand{\thetable}{S\arabic{table}}
\renewcommand{\thefigure}{S\arabic{figure}}
\renewcommand{\theHtheorem}{S\arabic{theorem}}
\renewcommand{\theHequation}{S\arabic{equation}}
\renewcommand{\theHtable}{S\arabic{table}}
\renewcommand{\theHfigure}{S\arabic{figure}}

\begin{center}
{\Large\bf Appendices\par}
\end{center}
\vspace{6pt}

\noindent These appendices contain the complete proofs, the algorithms and
software description, the full protocol and counterevidence for the
language-model score-bank audit, the synthetic and comparator validation, the
prespecified validation protocols (the essential study E1, which has not
been run; the vision study O1; and the auxiliary language replication E2),
and a brief statement of scope and limitations. Unprefixed theorem, equation, table,
figure and section numbers refer to the main paper; numbers prefixed with
``S'' and lettered sections refer to these appendices.

\sloppy
\input{sec/supp_proofs}
\input{sec/supp_computation}
\input{sec/supp_empirical_details}
\input{sec/supp_validation}
\input{sec/supp_protocols}
\input{sec/supp_limitations_appendix}
\end{document}

%% file: preamble.tex
\usepackage[T1]{fontenc}
\usepackage{amsthm,mathtools}
\usepackage{multirow,array}
\usepackage{url}
\usepackage{tikz}
\usetikzlibrary{arrows.meta}

\newtheorem{theorem}{Theorem}
\newtheorem{proposition}[theorem]{Proposition}

\newtheorem{lemma}[theorem]{Lemma}
\theoremstyle{definition}

\newcommand{\E}{\mathbb E}
\newcommand{\Prob}{\mathbb P}
\newcommand{\one}{\mathbf 1}
\newcommand{\AUC}{\operatorname{AUC}}
\newcommand{\TV}{\operatorname{TV}}
\newcommand{\sym}{\operatorname{sym}}
\newcommand{\supp}{\operatorname{supp}}
\newcommand{\DeltaK}{\Delta_K}
\newcommand{\Q}{\mathcal Q}
\newcommand{\ind}[1]{\mathbf 1\{#1\}}

\newcommand{\method}{\textsc{MixAudit}}

\newcommand{\sref}[1]{Appendix~\ref{#1}}
\newcommand{\mref}[1]{\ref{#1}}
\newcommand{\meqref}[1]{(\ref{#1})}

%% file: sec/0_abstract.tex
\begin{abstract}
Does an AUROC comparison survive a change in the proportions of tasks,
populations, or image corruptions? When failure detectors or uncertainty
scores are ranked by AUROC on a benchmark pooling heterogeneous groups, the
ranking describes one particular mix. We study
two fixed scores for a common binary target under mixtures of fixed
group-conditional laws. Building on the established cross-group AUC
decomposition, we prove that every rational symmetric quadratic comparison
geometry is realizable by finite scalar-score distributions. Universal
dominance is consequently coNP-hard, and even exhaustive checks of every
mixture on fewer than $K$ groups can miss a $K$-group reversal. A
grid bound gives a dimension-uniform approximation certificate
with sharp $1/m$ order. Common score marginals across groups make the
comparison numerator linear, with a quantitative error envelope for imperfect
alignment. These results yield exact finite-bank audits, explicit reversal
mixtures and distances, and simultaneous population statements under
independent sampling. For failure detection with a frozen ResNet-50 on
ImageNet-C, 15 of 18 confidence-score comparisons win in every corruption
family on a discovery split, and all 15 are finite-bank certified over the continuum of
family mixtures on a held-out split; over all 15 corruption
types, of 14 discovery winners the corrected grid certifies 10, leaves 3 unresolved, and finds one explicit
confirmation reversal of at least 0.05 AUROC. A complete
audit of 11,019 released Stable LM 2 12B outputs and 35 uncertainty methods
finds 66 confirmation-set comparisons that win within all four tasks yet
reverse under mixing. Discovery-fitted marginal rank alignment removes all
66, while an alignment-error envelope certifies 83 comparisons. Label checks
also change several selected reversals. The findings support a precise
distinction between robustness of frozen scores and robustness after
calibration; they do not establish human-error detection, robustness to
unseen corruptions, or generalization beyond the evaluated banks.
\end{abstract}

%% file: sec/1_intro.tex
\section{Introduction}
\label{sec:intro}
\input{figures/fig1}
An AUROC leaderboard answers a question about a particular population.
Deployment can change the proportions of tasks, sites, user groups, or image
corruptions while leaving each group's joint distribution of labels and
scores unchanged. A practical question is then: does the observed ordering of
two scores persist throughout the allowed changes in that mixture?

The question arises directly in failure detection and uncertainty estimation
for vision models. A confidence score for an image classifier, such as the
maximum softmax probability, is judged by how well it ranks misclassified
images above correctly classified ones. On a corruption benchmark such as
ImageNet-C \citep{hendrycks2019imagenetc}, the pooled AUROC of such a score is
computed over a fixed mix of corruption types and severities, and a deployed
camera does not see that mix. The same holds for uncertainty scores of
language models evaluated on a fixed mix of tasks.

This question differs from reporting a high overall AUROC or a favorable AUROC
in every group. AUROC compares a positive example to an independently drawn
negative example, so cross-group comparisons contribute to the pooled metric.
The xAUC decomposition already makes this explicit
\citep{kallus2019xauc}. Recent work further studies AUC non-collapsibility,
cluster decomposition, and failure-mixture effects
\citep{matos2026collapsibility,sudjianto2025decomposing,cieslak2026audits}.
Our starting point is this established decomposition. Our question is a
\emph{universal comparison}: what evidence suffices to certify an ordering over
a continuum of deployment mixtures, and what can simpler checks miss?

We give two answers that materially constrain such audits. First, any rational
symmetric quadratic form, after a positive rescaling, is the AUROC difference
of two realizable scalar scores under group-mixture shift, even when every group
has positive-label probability $1/2$. Thus the apparent difficulty of a
quadratic comparison is not an artifact of allowing matrices that no scores
could produce: universal dominance is coNP-hard. Second, for every $m<K$, two
scores can be ordered correctly by \emph{every} mixture supported on at most
$m$ groups, yet reverse on the uniform mixture of all $K$ groups. In this
construction, one score has perfect AUROC inside every group. Neither
within-group checks nor exhaustive low-order mixture checks establish universal
dominance.

We then build \method, an audit for a specified mixture set. It returns a
valid lower bound, an explicit reversing mixture when found, and an unresolved
status when the statistical or computational evidence is insufficient. The
audit uses exact rational optimization for small group counts and simultaneous
paired U-statistic intervals for population statements. The statistical
intervals remain valid when the mixture is selected using the same evaluation
data. We also quantify how far a reference mixture can move before a specified
reversal becomes possible. A corrected finite grid gives a
dimension-independent approximation bound for the AUROC gap with optimal $1/m$
order. Aligning each method's score marginals across groups yields a tractable
comparison, with an explicit bound for departures from exact alignment.

The theory does not assert that a reversal is common or that an arbitrary
mixture describes a realistic deployment. We evaluate the audit in two
domains. In vision, we audit four confidence scores of a frozen ResNet-50 for
its top-1 errors on ImageNet-C, with corruption families ($K=4$) or all 15
corruption types ($K=15$) as groups. In language, we audit a released
Stable LM 2 12B score bank containing 11,019 question-answering outputs and 35
uncertainty methods, with all 595 pairs included. Exact results for these
finite banks are distinguished from population confidence statements.
\Cref{fig:teaser} summarizes the problem, the audit, and the main results.

\paragraph{Scope and related work.}
The quadratic AUC decomposition, concentration tools, and global-optimization
primitives are established tools. We contribute the scalar-score realizability
reduction, the hierarchy of insufficient mixture checks, a sharp ratio-specific
grid bound, and a comparison identity with an error bound for marginal
alignment. Their integration yields a reproducible comparison audit. We do not
propose a new uncertainty estimator or failure detector, or claim a first
certificate for shifted AUROC. \citet{weber2022certifying} already certify AUC
under Hellinger drift; \citet{li2023transportauc} study transport to a
specified target population; distributionally robust AUC training addresses a
different optimization problem \citep{ma2020drauc,dai2023drauc}. Our structure
is a fixed finite collection of group distributions with uncertain mixture
weights, which permits sharp finite-bank conclusions and exposes the limits of
subgroup-only validation.
Groupwise quantile maps and transport-based post-processing already serve
fairness and ranking objectives \citep{nandy2022postprocessing,liu2025fairpot}.
Covariate-adjusted ROC analysis has long used negative-class placement values
\citep{janes2009adjusting}. We study the universal \emph{comparison} induced
by fixed score representations, including the consequences of applying those
established transformations. Our aligned identity and approximation envelope
describe when within-group comparisons suffice; they do not prescribe a
fairness objective or replace a deployment-specific utility analysis.

%% file: figures/fig1.tex
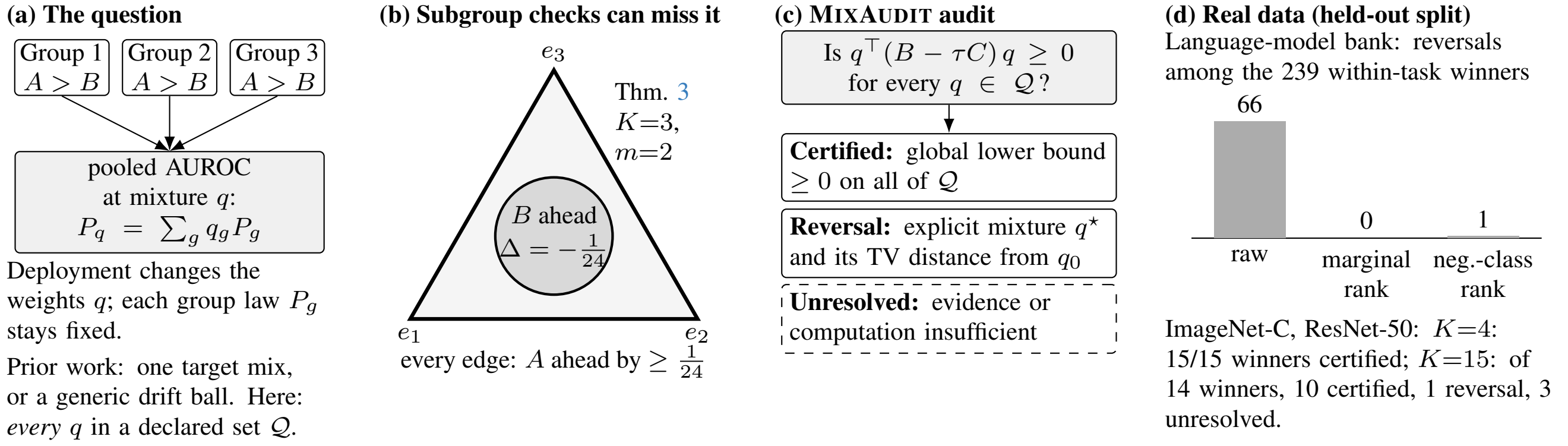
\begin{figure*}[t]
\centering
\footnotesize
\begin{tikzpicture}[
  font=\footnotesize,
  >={Latex[length=1.6mm]},
  ttl/.style={font=\footnotesize\bfseries, anchor=north west, inner sep=0pt},
  grp/.style={draw, rounded corners=1.5pt, minimum width=1.05cm, minimum height=0.62cm, align=center, inner sep=1pt, fill=white},
  outc/.style={draw, rounded corners=1.5pt, text width=3.55cm, align=left, inner sep=2.5pt, minimum height=0.62cm},
]
\begin{scope}
\node[ttl] at (0,4.05) {(a) The question};
\node[grp] (g1) at (0.62,3.35) {Group 1\\$A>B$};
\node[grp] (g2) at (1.82,3.35) {Group 2\\$A>B$};
\node[grp] (g3) at (3.02,3.35) {Group 3\\$A>B$};
\node[draw, fill=gray!12, rounded corners=1.5pt, text width=3.3cm, align=center, inner sep=2pt] (pool) at (1.82,1.85)
  {pooled AUROC at mixture $q$:\\ $P_q=\sum_g q_g P_g$};
\foreach \g in {g1,g2,g3} \draw[->] (\g.south) -- (pool.north);
\node[align=left, anchor=north west, text width=3.65cm, inner sep=0pt] at (0,1.2)
  {Deployment changes the weights $q$; each group law $P_g$ stays fixed.\\[2pt]
   Prior work: one target mix, or a generic drift ball. Here: \emph{every} $q$ in a declared set $\mathcal Q$.};
\end{scope}
\begin{scope}[xshift=4.15cm]
\node[ttl] at (0,4.05) {(b) Subgroup checks can miss it};
\coordinate (v1) at (0.35,0.55);
\coordinate (v2) at (3.55,0.55);
\coordinate (v3) at (1.95,3.321);
\fill[gray!8] (v1) -- (v2) -- (v3) -- cycle;
\fill[gray!30] (1.95,1.4738) circle (0.6532);
\draw[thick] (1.95,1.4738) circle (0.6532);
\draw[very thick] (v1) -- (v2) -- (v3) -- cycle;
\node[font=\footnotesize, align=center, inner sep=0pt] at (1.95,1.47) {$B$ ahead\\[1pt]$\Delta=-\frac{1}{24}$};
\node[anchor=north] at (v1) {$e_1$};
\node[anchor=north] at (v2) {$e_2$};
\node[anchor=south] at (v3) {$e_3$};
\node[anchor=north, align=center] at (1.95,0.36) {every edge: $A$ ahead by $\ge\frac{1}{24}$};
\node[anchor=west, align=left, text width=1.25cm, inner sep=0pt] at (2.62,2.75) {Thm.~\ref{thm:hierarchy}\\$K{=}3$, $m{=}2$};
\end{scope}
\begin{scope}[xshift=8.55cm]
\node[ttl] at (0,4.05) {(c) \method\ audit};
\node[draw, fill=gray!12, rounded corners=1.5pt, text width=3.55cm, align=center, inner sep=2.5pt] (q) at (1.95,3.35)
  {Is $q^\top(B-\tau C)\,q\ge0$\\ for every $q\in\mathcal Q$\,?};
\node[outc, anchor=north] (c1) at (1.95,2.62) {\textbf{Certified:} global lower bound $\ge0$ on all of $\mathcal Q$};
\node[outc, anchor=north] (c2) at (1.95,1.78) {\textbf{Reversal:} explicit mixture $q^\star$ and its TV distance from $q_0$};
\node[outc, anchor=north, dashed] (c3) at (1.95,0.94) {\textbf{Unresolved:} evidence or computation insufficient};
\draw[->] (q.south) -- (c1.north);
\end{scope}
\begin{scope}[xshift=12.9cm]
\node[ttl] at (0,4.05) {(d) Real data (held-out split)};
\node[anchor=north west, inner sep=0pt, text width=4.45cm] at (0,3.75)
  {Language-model bank: reversals among the 239 within-task winners};
\draw (0.3,1.45) -- (4.2,1.45);
\fill[gray!65] (0.55,1.45) rectangle (1.35,2.75);
\fill[gray!65] (3.15,1.45) rectangle (3.95,1.47);
\node[anchor=south] at (0.95,2.75) {66};
\node[anchor=south] at (2.25,1.45) {0};
\node[anchor=south] at (3.55,1.47) {1};
\node[anchor=north, align=center] at (0.95,1.42) {raw};
\node[anchor=north, align=center] at (2.25,1.42) {marginal\\rank};
\node[anchor=north, align=center] at (3.55,1.42) {neg.-class\\rank};
\node[anchor=north west, inner sep=0pt, text width=4.45cm, align=left] at (0,0.55)
  {ImageNet-C, ResNet-50: $K{=}4$: 15/15 winners certified; $K{=}15$: of 14 winners, 10 certified, 1 reversal, 3 unresolved.};
\end{scope}
\end{tikzpicture}
\caption{Overview. (a)~Pooled AUROC ranks two scores $A$ and $B$ for one target (\eg, two failure detectors
for one classifier) on one mixture of groups such as tasks or corruption types. We ask whether the ordering holds
for every mixture in a declared set, not for one target population or a generic drift ball. (b)~In the realizable
construction of Theorem~\ref{thm:hierarchy} ($c=17/24$), $A$ wins in every group and on every two-group mixture, yet $B$
wins on the shaded disc. (c)~\method\ returns a certificate, an explicit reversing mixture with its distance, or
``unresolved''. (d)~Finite-bank results on held-out data: marginal rank alignment removes all 66 language-model
reversals (Table~\ref{tab:calibration}); vision results are in Table~\ref{tab:o1}.}
\label{fig:teaser}
\end{figure*}

%% file: sec/2_formulation.tex
\section{Comparison Under Group-Mixture Shift}
\label{sec:formulation}
Let $G\in\{1,\ldots,K\}$ denote a prespecified group and let
$Z=(Y,S_A,S_B)$, where $Y\in\{0,1\}$ is one common binary target and larger
scores indicate $Y=1$. The two scores can be classifiers for the same outcome
or uncertainty estimators for the same base model's errors,
such as failure detectors for one image classifier. Comparing error
detectors attached to different base models generally changes $Y$ and is
outside this formulation.

The conditional laws $P_g=\mathcal L(Z\mid G=g)$ are fixed. A deployment
mixture $q\in\DeltaK=\{q\ge0:\one^\top q=1\}$ defines
$P_q=\sum_g q_gP_g$. Write $\pi_g=\Prob_g(Y=1)$ and
$p(q)=\pi^\top q$. We assume $0<\pi_g<1$ for the universal simplex statements;
more generally restrict $\Q$ to mixtures with $0<p(q)<1$. No conclusion covers
within-group distribution shift unless it is included separately in the audit.

Let $\phi(s,t)=\ind{s>t}+\tfrac12\ind{s=t}$, with independent draws for
the positive and negative observations. The cross-group comparison difference is
\begin{equation}
 d_{gh}=\E[\phi(S_A^g,S_A^h)-\phi(S_B^g,S_B^h)
             \mid Y_g=1,Y_h=0].
\end{equation}
Define symmetric matrices
\begin{align}
 B&=\sym\!\left(\operatorname{diag}(\pi)\,
 d\,\operatorname{diag}(\one-\pi)\right),\label{eq:B}\\
 C&=\tfrac12(\pi\one^\top+\one\pi^\top)-\pi\pi^\top.\label{eq:C}
\end{align}
Here $\sym(M)=(M+M^\top)/2$. The familiar cross-group decomposition gives
the following useful representation.
\begin{proposition}[Comparison identity]
\label{prop:identity}
For any admissible mixture,
\begin{equation}
 \begin{aligned}
 \Delta(q)&:=\AUC_A(q)-\AUC_B(q)=\frac{q^\top Bq}{q^\top Cq},\\
 q^\top Cq&=p(q)(1-p(q))>0.
 \end{aligned}\label{eq:ratio}
\end{equation}
Consequently $\Delta(q)\ge\tau$ for all $q\in\Q$ if and only if
$\min_{q\in\Q}q^\top(B-\tau C)q\ge0$.
\end{proposition}
A proof is included in \sref{app:proofs} for completeness. On the full simplex, the condition for
$\tau=0$ is precisely copositivity of $B$: $x^\top Bx\ge0$ for all $x\ge0$.
For two groups it reduces to
\begin{equation}
 B_{11}\ge0,\quad B_{22}\ge0,\quad
 B_{12}+\sqrt{B_{11}B_{22}}\ge0.\label{eq:two}
\end{equation}
The off-diagonal condition explains why two positive within-group gaps do not
suffice.

\paragraph{Why the common mixture matters.}
The positive and negative group weights are coupled:
$r_g^+(q)=q_g\pi_g/p(q)$ and
$r_g^-(q)=q_g(1-\pi_g)/(1-p(q))$.
If these two class-conditional weights could instead vary \emph{independently}
over their simplices, the worst gap would simply be $\min_{g,h}d_{gh}$.
Universal dominance in that larger shift class is equivalent to entrywise
nonnegativity of $d$. The coupled shift studied here admits more comparisons
as robust, but has the harder geometry characterized below.

\paragraph{The estimand must be declared.}
A macro-average of within-task AUROCs is a different metric: it intentionally
omits cross-task comparisons. It is appropriate when rankings are used only
within each task. Pooled AUROC and this audit apply when a common score ranks
examples across tasks or populations. A separate monotone transformation in
each group can change that pooled ordering; a common strictly increasing
transformation of one method cannot. We freeze score orientation, groups,
labels, and any calibration before evaluation.

\paragraph{Allowed mixtures.}
We support a rational polytope $\Q\subseteq\DeltaK$, including group-share
bounds and $\TV(q,q_0)=\tfrac12\|q-q_0\|_1\le r$. The full simplex is a
stress test. A deployment claim must use an externally justified $\Q$;
choosing a convenient region after observing the results changes the claim.

%% file: sec/3_limits.tex
\section{What Subgroup Checks Cannot Establish}
\label{sec:limits}
The matrix representation alone leaves a possible escape: perhaps matrices
generated by real scalar scores form an easy subclass. They do not.

\begin{theorem}[Realizable quadratic comparisons]
\label{thm:realize}
For every rational symmetric $K\times K$ matrix $D$ with $|D_{gh}|\le1$,
there exist finite-support joint distributions of two scalar scores and one
binary label, with $\pi_g=1/2$ for every group, such that
\begin{equation}
 \Delta(q)=\varepsilon q^\top Dq\ (q\in\DeltaK),
 \qquad \varepsilon=\frac{1}{4(K+1)}.
\end{equation}
The distributions have a description of polynomial size in the input. It is
therefore coNP-hard to decide whether one score weakly dominates another for
every mixture, even under equal group prevalences and finite scalar-score
supports.
\end{theorem}

\paragraph{Construction.}
In negative group $h$, both scores equal $h$. Baseline positive scores are
uniform on $\{1/2,3/2,\ldots,K+1/2\}$. In positive group $g$, perturb the
other score's tail probability at threshold $h$ by $\varepsilon D_{gh}$.
Telescoping differences define nonnegative probabilities: each interior mass
changes by at most $2\varepsilon$, less than its baseline mass. The cross-AUC
differences are exactly the prescribed perturbations. \sref{app:proofs}
verifies realizability, bit complexity, and the reduction from matrix
copositivity \citep{murty1987complexity}. This is an exact worst-case complexity
statement; it does not claim that every practical small-$K$ audit is hard or
establish hardness at a fixed, dimension-independent approximation margin.

\begin{theorem}[An arbitrary hierarchy of missed reversals]
\label{thm:hierarchy}
For any integers $1\le m<K$, there exist two scores, with equal prevalence
$1/2$ in every group, such that $\AUC_A(e_g)=1$ for every $g$, and
\begin{equation}
 \begin{aligned}
 \Delta(q)&>0\quad\text{whenever }|\supp(q)|\le m,\\
 \Delta(\one/K)&<0.
 \end{aligned}
\end{equation}
The construction can have
\begin{equation}
 \AUC_A(q)=\tfrac12+\tfrac12\|q\|_2^2,
 \qquad\AUC_B(q)=c,
\end{equation}
for any $\tfrac12+\tfrac1{2K}<c<\tfrac12+\tfrac1{2m}$.
\end{theorem}

Here $S_A=4G+Y$ is perfect within groups but orders different groups almost
entirely by group identity. For $S_B$, all negatives have score zero and each
positive has score $1$ with probability $c$, otherwise $-1$, independently of
its group. The gap is controlled by the concentration of $q$. For
$c=\tfrac12+\tfrac14(K^{-1}+m^{-1})$, both the smallest positive gap on
all $m$-group mixtures and the magnitude of the negative uniform gap equal
$\tfrac14(m^{-1}-K^{-1})$. This supplies an explicit quantitative example,
including the case $m=K-1$; \cref{fig:teaser}(b) shows $K=3$, $m=2$.

These examples do not show that subgroup reporting is useless. They show
precisely which additional claim it cannot support. A continuum audit must
account for the interactions between groups; isolated group wins, all
two-group wins, and a favorable reference mixture are insufficient in general.

%% file: sec/4_audit.tex
\section{A Certified Comparison Audit}
\label{sec:audit}
\subsection{Exact Finite-Bank Geometry}
For a released evaluation bank, define $P_g$ to be the empirical distribution
of its rows. Pairwise win counts, including half credit for ties, give rational
$B$ and $C$ exactly. This is a fully specified finite-distribution estimand;
optimizing it introduces no sampling assertion. \method\ minimizes
$q^\top Dq$ on $\Q$, with $D=B-\tau C$ for a desired AUROC margin $\tau$.

For a cell $\mathcal P\subseteq\Q$ with vertices $v_1,\ldots,v_J$, write
$q=\sum_i\alpha_i v_i$. Then
\begin{equation}
 \min_{i,j} v_i^\top Dv_j
 \ \le\ \min_{q\in\mathcal P}q^\top Dq
 \ \le\ \min_i v_i^\top Dv_i.\label{eq:cellbound}
\end{equation}
Subdivision tightens these bounds. Maintaining cells that cover the feasible
set, every completed iteration returns a global lower
bound and a feasible upper-bound witness; an uncorrected grid or local optimizer
cannot supply the former. Rational arithmetic avoids interpreting numerical
tolerances as proof. For small $K$, enumerating stationary points on all faces
gives an exact fallback, including zero-margin cases. These are classical
quadratic-optimization and copositivity techniques
\citep{bundfuss2008simplicial,bundfuss2009adaptive}, adapted here to the
comparison estimand. Their cost can grow exponentially; we apply them
only with $K\le4$.

\begin{theorem}[A corrected grid certificate]
\label{thm:grid}
Let $m\ge2$, $\mathcal G_m=\{a/m:a\in\mathbb N_0^K,\sum_g a_g=m\}$,
$\gamma_m=\min_{q\in\mathcal G_m}\Delta(q)$,
$\gamma=\min_{q\in\DeltaK}\Delta(q)$, and
$b=\max_g\Delta(e_g)$. For arbitrary group prevalences in $(0,1)$,
\begin{equation}
 \frac{m\gamma_m-b}{m-1}\le\gamma\le\gamma_m.\label{eq:gridbound}
\end{equation}
The $1/m$ correction has optimal order: realizable scores with $K\ge2m$
can satisfy $\gamma_m-\gamma\ge1/(4m)$.
\end{theorem}
This specializes the classical random-walk argument for simplex optimization
\citep{nesterov2003random} to the normalized AUROC ratio. It is valid without
an inverse-minimum-prevalence factor. Thus a grid \emph{with} its correction
can certify dominance, while an uncorrected positive grid minimum cannot.
There are $\binom{K+m-1}{m}$ grid points: fixed-accuracy approximation and
exact weak dominance have different complexity. The support hierarchy is
consistent with this bound; its missed reversals shrink at order $1/m$.

\paragraph{Reversal distance.}
For a reference mixture $q_0$ and a meaningful gap $\kappa>0$, define
\begin{equation}
 r_\kappa=\inf\{\TV(q,q_0):q\in\Q,\ \Delta(q)\le-\kappa\},\label{eq:radius}
\end{equation}
with $r_\kappa=+\infty$ if the set is empty. Bisection on the radius and
certified minimization of $q^\top(B+\kappa C)q$ yields a bracket. A feasible
reversal gives an upper bound; a strictly positive global lower bound on the
smaller ball gives a lower bound. We use $\kappa=0.01$ for a practically
interpretable sensitivity analysis. Positive-margin reversal avoids conflating a numerically tiny tie
with a material ranking change.

\subsection{Population Statements After Mixture Selection}
For population inference, suppose the audit receives $n_g\ge2$ independent
iid observations from each fixed group law, independently of model fitting
and protocol choices. Define
\begin{align}
 h(z,z')&=y(1-y')\{\phi(s_A,s'_A)-\phi(s_B,s'_B)\},\nonumber\\
 k(z,z')&=\tfrac12\{h(z,z')+h(z',z)\}.
 \label{eq:kernel}
\end{align}
The kernel is symmetric, has range $[-1/2,1/2]$, and satisfies
$B_{gh}=\E[k(Z_g,Z'_h)]$. Estimate off-diagonal entries by all cross-group
pairs and diagonal entries by unordered pairs of \emph{distinct} observations.
The diagonal is an unbiased U-statistic, which differs from the empirical-bank
product estimator by a factor $n_g/(n_g-1)$.

\begin{theorem}[A simultaneous comparison envelope]
\label{thm:bands}
Let $J$ score comparisons be fixed before evaluation,
$L=JK(K+1)/2$, $m_{gh}=\min(n_g,n_h)$ for $g\ne h$, and
$m_{gg}=\lfloor n_g/2\rfloor$. For any $\delta\in(0,1)$, set
\begin{equation}
 e_{gh}=\sqrt{\frac{\log(2L/\delta)}{2m_{gh}}}.
\end{equation}
With probability at least $1-\delta$, simultaneously for all $J$ comparisons
and every $q\in\DeltaK$,
\begin{equation}
 q^\top(\widehat B-E)q\le q^\top Bq
        \le q^\top(\widehat B+E)q.\label{eq:uniform}
\end{equation}
Thus the same statement holds for any data-selected mixture and mixture set.
It does not justify data-selected score training, group definitions, or labels
outside the prespecified family.
\end{theorem}

The proof uses permutation blocking for the U-statistics and a union bound;
it does not treat the quadratically many pairs as independent observations.
Clipping entry intervals to $[-1/2,1/2]$ preserves coverage. The implementation
also intersects these bands with empirical-Bernstein intervals from
independent, fixed matched pairs \citep{maurer2009empirical}, dividing the
error budget between the two constructions. This can exploit correlation
between the competing scores. Complete formulas and the extension to
independent clusters are in \sref{app:statistics}.

If $\underline B\le B\le\overline B$ entrywise, a nonnegative certified
minimum of $q^\top\underline Bq$ proves dominance throughout $\Q$ on the
coverage event. A feasible $q$ with $q^\top\overline Bq<0$ proves a reversal
at that mixture. Otherwise the result is \emph{unresolved}. This three-way
rule separates an actual ordering from insufficient evidence. Prevalence
estimation is unnecessary for the sign; quantitative AUROC-gap intervals
add simultaneous prevalence bounds and retain the positive denominator.

The audit also reports within-group, macro, and reference-mixture AUROCs
as descriptive baselines; they are not substituted for a continuum
certificate.

\subsection{A Tractable Boundary: Aligning Score Distributions}
\label{sec:alignment}
Calibration changes the frozen score and therefore changes the comparison
problem. A useful boundary occurs when each method has the same marginal
score distribution in every group. The two methods need not share a marginal.
Let $a_g=\pi_g(1-\pi_g)\Delta(e_g)$.
\begin{theorem}[Comparison after marginal alignment]
\label{thm:alignment}
If $\mathcal L(S_s\mid G=g)$ does not depend on $g$ for each $s\in\{A,B\}$,
then, including scores with ties,
\begin{equation}
 \Delta(q)=\frac{\sum_g q_ga_g}{p(q)(1-p(q))}.\label{eq:aligned}
\end{equation}
Universal dominance on the full simplex is equivalent to nonnegative
within-group gaps. More generally let
$\delta_{s,gh}=\|F_{s,g}^{\rm mid}-F_{s,h}^{\rm mid}\|_\infty$ and
$w_{gh}=\min\{\pi_g+\pi_h,2-\pi_g-\pi_h\}$. Then
\begin{align}
 \left|p(q)(1-p(q))\Delta(q)-q^\top a\right|
 &\le\sum_{g<h}q_gq_hw_{gh}\nonumber\\[-3pt]
 &\quad\cdot(\delta_{A,gh}+\delta_{B,gh}).\label{eq:alignmenterror}
\end{align}
\end{theorem}
Here $F^{\rm mid}(t)=\Prob(S<t)+\tfrac12\Prob(S=t)$. The exact identity
uses the AUC placement-value representation. Groupwise randomized
probability-integral transforms align marginals and preserve within-group
AUROCs in expectation, using unlabeled calibration data in the population
limit. Quantile alignment and demographic-parity transport are established
methods \citep{jiang2020wasserstein}; negative-class placement calibration is
also classical \citep{janes2009adjusting}. We contribute the comparison
identity and its error bound, not a new calibration algorithm.

Alignment changes cross-group rankings, can discard useful group risk
information, and estimated transforms need not align new data exactly. The
completed calibration control below measures this boundary rather than
assuming that raw-score reversals survive preprocessing.

%% file: sec/5_experiments.tex
\section{Experiments}
\label{sec:experiments}
We report two completed real-data studies, a released language-model score
bank (\cref{sec:llm,sec:calibration}) and a vision failure-detection study
(\cref{sec:vision}), and a synthetic coverage study (\cref{sec:coverage}).
\begin{table*}[t]
\centering\small
\caption{All five discovery-selected examples, evaluated on confirmation.
The two methods and the stress mixture remain fixed across label checks.
Positive values favor the first method. ``Task check'' uses the prespecified
task-dependent scoring rules in \sref{app:empirical-details}; it is not human
adjudication. Spectral and density names are abbreviated only in this table.}
\label{tab:examples}
\begin{tabular}{lrrrrr}
\toprule
Pair $A-B$ & Reference & \shortstack{Minimum\\within} & \shortstack{Stress:\\released} & \shortstack{Stress:\\task check} & \shortstack{Stress:\\AlignScore}\\
\midrule
Token entropy--Mahalanobis & $+.15085$ & $+.08707$ & $-.12342$ & $+.21626$ & $+.30665$\\
Sequence probability--NLI degree & $+.07991$ & $+.01424$ & $-.07932$ & $-.07592$ & $-.04745$\\
ROUGE-L--normalized entropy & $+.01628$ & $+.03048$ & $-.10170$ & $+.09561$ & $+.14390$\\
CCP--NLI eccentricity & $+.03952$ & $+.01298$ & $-.06089$ & $-.05539$ & $-.04968$\\
NLI eccentricity--Jaccard eigenvalue & $-.01490$ & $-.00809$ & $-.11478$ & $+.00920$ & $+.07352$\\
\bottomrule
\end{tabular}
\end{table*}
\begin{figure*}[t]
\centering
\includegraphics[width=\textwidth]{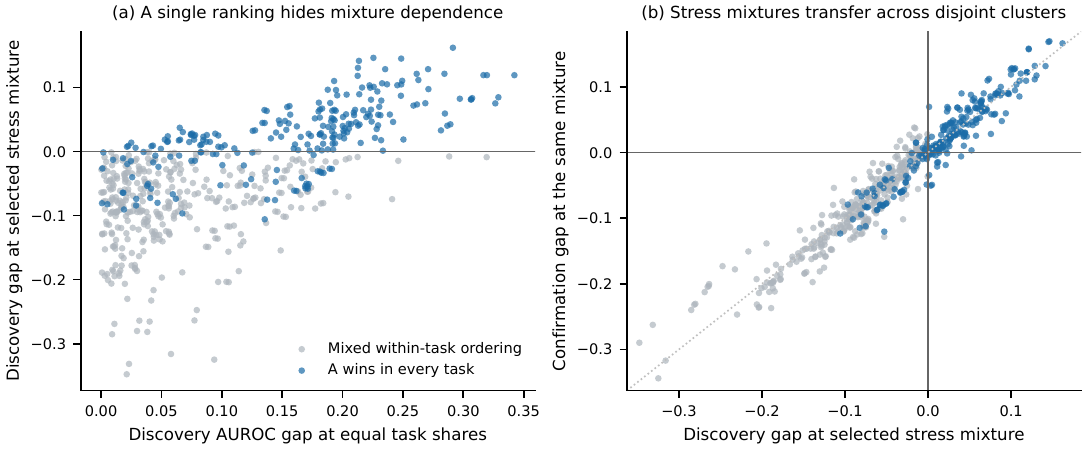}
\caption{All 595 primary-endpoint pairs. Blue points denote pairs where the
discovery reference winner wins within each task; gray points have mixed
within-task ordering. Left: equal-task-reference advantage versus gap at the
selected discovery stress mixture. Right: the same stress mixtures evaluated
on disjoint confirmation clusters. The stress grid supplies feasible gaps;
continuous rational optimization independently establishes the reversal
counts. These are finite-bank comparisons, not population confidence plots.}
\label{fig:all-pair-audit}
\end{figure*}

\begin{figure*}[t]
\centering
\includegraphics[width=\textwidth]{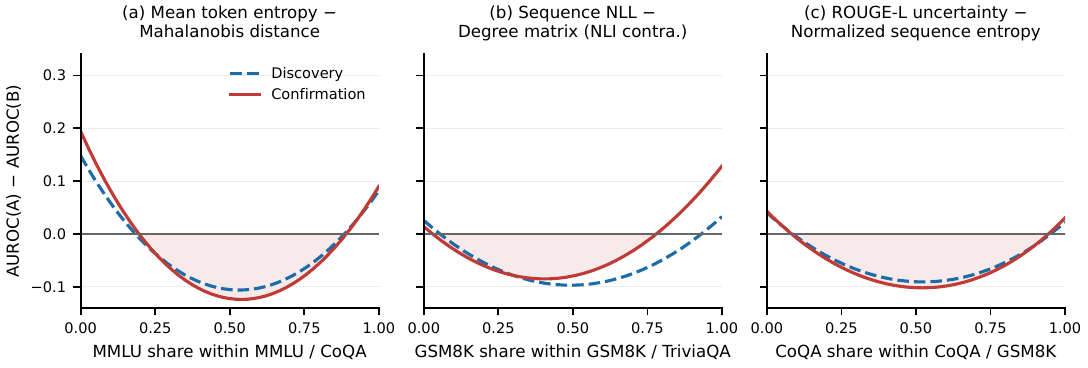}
\caption{The first three illustrations under the frozen discovery-only
selection rule. Only the two displayed task proportions vary; the other
two task masses are zero. Each discovery-selected method $A$ wins in all
four individual tasks, yet the intermediate mixtures can favor $B$.
Lines evaluate the exact finite-bank AUROC function on a dense display
grid, using the released Accuracy endpoint. Confirmation uses disjoint
input/story clusters. These are not population-confidence curves. The
first and third examples change sign under the alternative correctness
checks in Table~\ref{tab:examples}; that label sensitivity is retained.}
\label{fig:fixed-mixture-curves}
\end{figure*}
\subsection{Released Language-Model Score Bank}
\label{sec:llm}
We reanalyze the released LM-Polygraph Stable LM 2 12B \emph{base-model}
results \citep{vashurin2025polygraph}: MMLU (5,700 rows), CoQA (2,000),
GSM8K (1,319), and TriviaQA (2,000). All 35 released scalar uncertainty
methods are finite on all 11,019 rows, so there is no selective deletion.
The original benchmark did not claim the pooled-mixture robustness studied
here; this is a new analysis of its saved outputs. No model inference is
represented as rerun.

Before computing new comparison outcomes, we froze all 595 unordered method
pairs, the primary released Accuracy label, score orientation, and a
deterministic cluster split. CoQA questions sharing a story remain together;
identical input strings in the other tasks remain together. Discovery and
confirmation contain 5,605 and 5,414 rows. This is a retrospectively specified
analysis protocol, not prospective preregistration. Each pair is oriented by
the discovery winner at $q_0=(1/4,1/4,1/4,1/4)$, with that orientation fixed
on confirmation. The comparison domains are the full simplex and the
restricted region $q_g\ge0.05$. Witnesses and five displayed pairs are
selected using discovery only; complete pair tables include every outcome.

\paragraph{Exact finite-bank results.}
On discovery, exact continuum optimization classifies 419 pairs as
mixture-dependent and 176 as universally ordered. Of the 238 pairs whose
discovery winner also wins in every individual task, 62 nevertheless reverse
under a mixture. At the unchanged discovery witnesses, 390 of the 419
reversals and 55 of those 62 persist on confirmation
(\cref{fig:all-pair-audit}). The restricted domain
still has 350 mixture-dependent discovery comparisons. These counts describe
this finite bank and comparison family, not a prevalence estimate for all
uncertainty methods or a population significance test.

The corrected denominator-40 grid certifies 168 of the 176 discovery-wide
orderings and finds all 419 negative witnesses; eight cases are unresolved
until exact optimization. On confirmation it certifies 156 of 173 orderings,
with 17 unresolved. Exhaustive two-group continuum checks find \emph{every}
raw-bank reversal here. Thus the hierarchy in \cref{thm:hierarchy} is
a worst-case theoretical limitation, not an additional four-way failure
observed in these data.

\paragraph{Labels and stability limit the interpretation.}
We reproduce all released Accuracy values from saved answers. In CoQA this
is literal stripped-string equality, which differs materially from a
semantic notion of correctness. Using the \emph{same} primary pairs and
mixtures, only 13 of the 62 discovery within-task-win reversals remain
negative on confirmation under all three label definitions. \Cref{tab:examples}
shows both stable reversals and sign changes, and
\cref{fig:fixed-mixture-curves} traces the first three pairs along two-task
mixtures. For the second example, a
one-AUROC-point reversal first becomes possible at discovery TV distance
$[.12988,.13086]$; the separately optimized confirmation radius is
$[.16602,.16699]$. The smallest-shift discovery witness has a positive
confirmation gap $+.00725$. A large-stress reversal can replicate even when
the exact estimated radius location does not.

\subsection{Calibration Is a Decisive Control}
\label{sec:calibration}
\begin{table*}[t]
\centering
\caption{Vision study (O1): ResNet-50 failure detection on ImageNet-C, 18
comparisons (six score pairs at three severities). Wins: all-group winners
on discovery. Cert.: confirmation continuum certificates. Rev.\ witness:
explicit confirmation reversal witnesses. Unres.: unresolved. Frozen rev.:
among discovery winners, reversals of the frozen discovery witness on
confirmation, material ones in parentheses; for $K=15$ only the material
frozen-grid count is computed (--).}
\label{tab:o1}
\input{generated/o1.tex}
\end{table*}
After the primary analysis, we froze a separate control that fits
group-specific marginal or negative-class rank transforms on discovery and
applies them to confirmation. This control is explicitly post hoc. Strict
monotone interpolation preserves all 140 within-task AUROCs exactly in the
saved bank; empirical step-CDF controls have only small additional tie effects.
\Cref{tab:calibration} compares the same confirmation rows and fixed pair
orientations. Marginal alignment removes all 66 reversals among the same 239
strict within-task winners. Overall mixture dependence remains because many
methods have different within-task advantages. Calibration is not a new
method here, and these results do not support a claim that the strongest raw
scale pathology persists after alignment.

\begin{table}[t]
\centering\small
\caption{Confirmation-bank control, all 595 pairs. ``Within-win reversals''
uses the same 239 strict all-task winners. ``Envelope'' counts universal
dominance certificates from the marginal-alignment error bound.}
\label{tab:calibration}
\setlength{\tabcolsep}{3pt}
\begin{tabular}{@{}lrrr@{}}
\toprule
Scores & Mixed pairs & Within-win & Envelope\\
 & & reversals & certificates\\
\midrule
Raw & 422 & 66 & 0\\
Marginal rank & 347 & 0 & 83\\
Negative-class rank & 348 & 1 & 0\\
\bottomrule
\end{tabular}
\end{table}

The median method's largest pairwise group KS distance falls from $.7856$
to $.0844$ after strict marginal alignment. The finite-bank envelope in
\cref{thm:alignment} certifies 83 comparisons using within-group
gaps and marginal discrepancies alone. All 47,600 checked matrix entries
across raw and four calibrated controls obey its bounds exactly. These are
deterministic checks on empirical laws, not sampling-coverage observations.
The implementation uses ordinary KS distances, which upper-bound the
mid-CDF distances required by the theorem. The negative-class row is
evaluated with this \emph{marginal} envelope; its zero count does not rule
out certificates tailored to negative-class alignment.

\subsection{Vision: Failure Detection Across Corruption Mixtures}
\label{sec:vision}
The vision study (O1) asks whether the audit transfers to a second domain
and to more groups. It was specified in advance as an optional extension
(\sref{app:o1protocol}); inference was run once. We use frozen torchvision
ResNet-50 \texttt{IMAGENET1K\_V2} weights on 1,000 ImageNet validation
images selected by a fixed hash, under the 15 standard ImageNet-C
corruption types at severities $1,3,5$ \citep{hendrycks2019imagenetc}:
45,000 corrupted-image evaluations from verified official archives. The
target is $Y=1$ when top-1 classification is wrong, so each score is a
failure detector for the same classifier. We compare negative maximum
softmax probability~\citep{hendrycks2017baseline}, softmax entropy, negative top-two logit margin, and
negative log-sum-exp of logits at temperature one (the energy score~\citep{liu2020energy}), giving six pairs at each
severity, 18 comparisons in all. The classifier is not refit. Image IDs are
split by a hash into 342 calibration, 316 discovery, and 342 confirmation
images, and all corrupted versions of one image stay together. Top-1 error
differs widely across groups (over all 1,000 images), from 24.7\% (brightness, severity 1) to
93.0\% (impulse noise, severity 5), which is the setting in which the mix
of corruptions can matter.

At each severity, the corruption types are the groups. We report the
prespecified noise, blur, weather, and digital family aggregation ($K=4$),
audited with exact stationary-face and branch-and-bound methods, and the
full $K=15$ analysis, audited only with the corrected grids $m=2,3,4$
(\cref{thm:grid}). Each comparison is oriented once by raw discovery AUROC and
kept fixed in both panels: raw scores and the marginal-rank control of
\cref{sec:alignment}. A reversal is material at 0.01 AUROC.

\Cref{tab:o1} gives the outcome. For the four-family setting, 15 of 18
comparisons are all-family winners on discovery. All 15 are certified as
continuum winners on confirmation, the three non-winners carry explicit
confirmation reversal witnesses, and no frozen discovery witness reverses.
Under image-cluster resampling (2,000 replicates; pointwise, descriptive, not
a population certificate) every raw-panel winner keeps its 95\% interval
above $-0.01$ at the frozen witness; in the marginal-rank panel 14 of 15 do.
For all 15 corruptions, 14 of 18 comparisons win every group on discovery;
of these 14, the corrected confirmation grids certify 10, leave 3 unresolved,
and give an explicit confirmation reversal for one (severity 5, negative
top-two margin against negative log-sum-exp; confirmation gap at most $-0.050$
at the witness mixture). Of the four non-winners, three carry reversal
witnesses and one is unresolved. No frozen discovery witness reverses
materially. Raw and marginal-rank controls give identical counts.

Unlike on the raw language bank, within-family wins are thus certified over
the continuum of family mixtures on the confirmation images (a finite-bank
statement), while with 15 groups winning every corruption type does not by
itself carry over to every mixture. This is one frozen model on one
benchmark and a transfer check of the audit, not support for the theorems.

\subsection{Sampling Coverage and Power}
\label{sec:coverage}
We simulate the $K=3$, $c=17/24$ hierarchy with truly independent samples,
500 repetitions per sample size, a fixed comparison, and total error
probability $.05$. Both the all-pairs Hoeffding bands and their
paired-variance intersection cover every matrix entry in all 500 repetitions
at each of $n_g=500,2000,8000,32000$. A count of 500/500 has a 95\% Wilson
interval $[.9924,1]$; this is compatible with conservatism rather than evidence
of exact 95\% calibration. At the known uniform reversing mixture, the
intersection certifies the negative sign in $0,0,32,500$ repetitions,
respectively; Hoeffding alone does so in $0,0,0,236$.
\sref{app:validation} gives all outcomes and the power figure. We do not apply these iid-row
intervals to correlated CoQA questions or describe the released bank as a
population-certified leaderboard.

\paragraph{Further validation.}
A prespecified replication with an independent language model and
human-audited correctness labels (E1) is specified in \sref{app:essential};
it has not been run, and no claim here depends on it. A smaller auxiliary
replication with Qwen3-1.7B (E2) was run and is reported in
\sref{app:e2results}; it does not replace E1.

%% file: generated/o1.tex
\begin{center}\small
\begin{tabular}{llrrrrr}
\toprule
Groups & Panel & Wins & \shortstack{Frozen rev.\\(material)} & Cert. & \shortstack{Rev.\\witness} & Unres.\\
\midrule
$K=4$ & Raw & 15/18 & 0 (0) & 15 & 3 & 0\\
$K=4$ & Marginal rank & 15/18 & 0 (0) & 15 & 3 & 0\\
$K=15$ & Raw & 14/18 & -- (0) & 10 & 4 & 4\\
$K=15$ & Marginal rank & 14/18 & -- (0) & 10 & 4 & 4\\
\bottomrule
\end{tabular}
\end{center}

%% file: sec/6_conclusion.tex
\section{Discussion and Conclusion}
A pooled AUROC ordering has a population-mixture domain. Within-group wins
and low-order stress tests cannot establish its universal validity in
general, even for realizable scalar scores. Corrected grids and exact small
group-count optimization can supply that missing certificate, and marginal
alignment identifies a tractable boundary that removes the cross-group scale
effect that otherwise makes the comparison difficult. On the language-model
bank, raw reversals among within-task winners occur but are sensitive to
labels and calibration. On ImageNet-C, the audit certifies the
corruption-family winners among four confidence scores of one frozen
ResNet-50, while over 15 corruption types it finds one winner that reverses
and leaves three unresolved.

\paragraph{Limitations.}
The empirical evidence covers two finite banks: a released language-model
bank whose labels are task-dependent string matches rather than validated
correctness, and one frozen classifier on one corruption benchmark. It does
not speak to unseen corruptions or other models, and the independent
replication (E1) has not been run. Every observed raw reversal was already
detectable on a two-group edge, so the higher-order hierarchy is a
worst-case result. Distribution-free population certificates need large
samples, and mixture certification does not address within-group shift,
fairness, or downstream decision utility.

%% file: sec/7_llm_statement.tex
\section*{LLM Usage Statement}
The authors developed the research ideas and proofs. AI tools provided coding, debugging, writing, and editing support, including polishing the exposition and presentation of the proofs. The authors reviewed and verified the resulting material and retain responsibility for the research questions, methods, proofs, experiments, interpretation, and final content.

%% file: sec/supp_proofs.tex
\section{Complete Proofs}
\label{app:proofs}
All scores are real-valued and measurable; their joint laws with the binary
target may be arbitrary. Independent draws, including the two draws used for
AUROC, are explicitly distinguished from correlated score coordinates on the
same example. Ties always receive half credit.

\subsection{Comparison Identity and the Two-Group Criterion}
\begin{proof}[Proof of Proposition~\mref{prop:identity}]
By Bayes' rule the positive-class group mass is $q_g\pi_g/p(q)$ and the
negative-class group mass is $q_h(1-\pi_h)/(1-p(q))$. Independence of the two
AUROC draws gives
\[
 \Delta(q)=\sum_{g,h}\frac{q_gq_h\pi_g(1-\pi_h)d_{gh}}
                         {p(q)(1-p(q))}.
\]
The quadratic form is unchanged by symmetrization. Since $\one^\top q=1$,
$q^\top Cq=(\pi^\top q)(\one^\top q)-(\pi^\top q)^2=p(q)(1-p(q))$.
The denominator is positive on the stated domain, so subtracting $\tau$ and
multiplying by it proves the margin equivalence. For the full simplex, every
nonzero nonnegative vector $x$ is a positive multiple of
$q=x/(\one^\top x)$; homogeneity proves the copositivity equivalence.
\end{proof}

For independently variable class-conditional group weights $r^+,r^-$, the
gap is $\sum_{g,h}r_g^+r_h^-d_{gh}$, a convex combination of entries of $d$.
Its minimum is the smallest entry, attained at corresponding point masses.
The common-$q$ coupling imposes a different feasible set of weights.

\begin{proof}[Proof of the two-group criterion, Eq.~\meqref{eq:two}]
Let $B=\left(\begin{smallmatrix}a&b\\b&c\end{smallmatrix}\right)$. Testing
coordinate vectors gives $a,c\ge0$. If $a,c>0$ but $b< -\sqrt{ac}$, the
vector $(\sqrt c,\sqrt a)$ gives a negative value. If $a=0,c>0,b<0$, use
$(t,1)$ for $t>c/(-2b)$; the symmetric case is identical. If $a=c=0,b<0$,
use $(1,1)$. This proves necessity including zero-diagonal cases. Conversely,
for nonnegative $u,v$,
\[
 au^2+2buv+cv^2=(\sqrt a\,u-\sqrt c\,v)^2+
                 2(b+\sqrt{ac})uv\ge0.
\]
For an exact edge minimum on $q=(t,1-t)$, evaluate the two endpoints and,
when $a-2b+c>0$, the stationary point
$t=(c-b)/(a-2b+c)$ if it is in $[0,1]$. Nonpositive curvature requires only
endpoints. With unequal prevalences this minimizes the numerator, not
necessarily the AUROC ratio.
\end{proof}

\subsection{Realizability and Computational Hardness}
\begin{proof}[Proof of Theorem~\mref{thm:realize}]
Set $u=1/(K+1)$ and $\varepsilon=1/[4(K+1)]$. In negative group $h$, both
scores equal the integer $h$. In positive group $g$, score $B$ is uniform
on $j+1/2$, $j=0,\ldots,K$, and score $A$ has the masses
\begin{align*}
 w_{g0}&=u-\varepsilon D_{g1},\\
 w_{gj}&=u+\varepsilon(D_{gj}-D_{g,j+1}),\qquad 1\le j<K,\\
 w_{gK}&=u+\varepsilon D_{gK}.
\end{align*}
The two endpoint masses are at least $3u/4$, and every interior mass is at
least $u/2$, because $|D_{gh}|\le1$. Their perturbations telescope to zero,
so the masses sum to one. There are no positive-negative score ties. A
positive score exceeds the negative score $h$ exactly for indices $j\ge h$,
and therefore
\[
 \Prob(S_A^g>h\mid Y=1)-\Prob(S_B^g>h\mid Y=1)
 =\sum_{j=h}^K(w_{gj}-u)=\varepsilon D_{gh}.
\]
The desired comparisons are thus genuine scalar-score comparisons.

For a joint realization, choose the two positive score coordinates independently
given $(G,Y)$, and define a feature $X=(G,U,V)$ with scorers $s_A(X)=U$ and
$s_B(X)=V$. On negative examples set $U=V=G$. The scorers do not take the
label as an argument. Give every group label probability $1/2$. There are
$K$ negative support points and at most $K(K+1)^2$ positive support points;
their rational probabilities have encoding length polynomial in the encoding
of $D$. The comparison matrix is $B=\varepsilon D/4$, and the denominator
is $1/4$, proving $\Delta(q)=\varepsilon q^\top Dq$.

For hardness, take an arbitrary rational symmetric copositivity instance $M$,
let $a=\max\{1,\max_{g,h}|M_{gh}|\}$, and apply the construction to $D=M/a$.
Positive scaling preserves copositivity, and the scaling and construction
have polynomial encoding size and can be computed in polynomial time.
Universal dominance holds if and only if $M$ is copositive. Since
copositivity testing is coNP-hard \citep{murty1987complexity}, the claimed
restricted dominance problem is coNP-hard. This reduction establishes exact
decision hardness, not a fixed-margin approximation lower bound.
\end{proof}

\subsection{Support Hierarchy and an Analytic Radius}
\begin{proof}[Proof of Theorem~\mref{thm:hierarchy}]
Let $\pi_g=1/2$. Set $S_A=4g+Y$ in group $g$. A positive and negative from
the same group are ranked correctly, and for different groups the positive
is ranked higher exactly when its group index is larger. Both
class-conditional group mixtures equal $q$, so
\[
 \AUC_A(q)=\sum_gq_g^2+\sum_{g>h}q_gq_h
          =\frac12+\frac12\sum_gq_g^2.
\]
Let $S_B=0$ on negatives and let its positive value be $1$ with probability
$c$, otherwise $-1$, independently of the group. Then $\AUC_B(q)=c$.
These assignments are realized jointly by the score-coordinate construction
used in the preceding proof. If $q$ has at most $m$ nonzero coordinates,
Cauchy--Schwarz gives $\sum_gq_g^2\ge1/m$. Choose
$1/2+1/(2K)<c<1/2+1/(2m)$. Every such sparse mixture has positive gap, whereas
the uniform $K$-group mixture has negative gap. At a pure group, $\AUC_A=1$.
Choosing the midpoint value of $c$ gives the quantitative gaps stated in the
main text.
\end{proof}

\begin{figure}[ht]
\centering
\includegraphics[width=.6\textwidth]{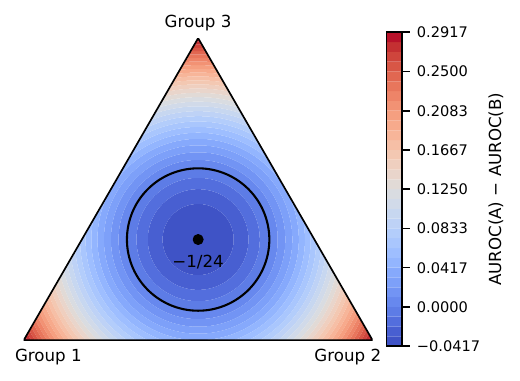}
\caption{Analytic construction in Theorem~\mref{thm:hierarchy}, $K=3$, $m=2$,
$c=17/24$. Every boundary mixture favors $A$ by at least $1/24$; the center
favors $B$ by $1/24$. Thus exhaustive checks along all three edges miss an
interior reversal. Color shows the AUROC difference.}
\label{fig:hierarchy}
\end{figure}

The construction also gives an exact radius oracle for implementation checks.
Take $q_0=e_1$ and a margin $\kappa>0$ such that
$t=2(c-\kappa)-1$ lies in $[1/K,1)$. At TV distance $r$ from $e_1$, one has
$q_1=1-r$, and the minimum possible squared norm is
$(1-r)^2+r^2/(K-1)$, attained by equal remaining coordinates. Thus
\begin{equation}
 r_\kappa=\frac{K-1}{K}\left(1-
       \sqrt{\frac{Kt-1}{K-1}}\right).\label{eq:analyticradius}
\end{equation}
If $t<1/K$, no such reversal exists; if $t\ge1$, the reference already meets
the target. The equality case $t=1/K$ is attained only at the uniform mixture.
Equation~\eqref{eq:analyticradius} is a test oracle for positive-margin
reversal, not an empirical result.

\subsection{A Sharp Corrected Grid Bound}
\begin{proof}[Proof of Theorem~\mref{thm:grid}]
The denominator is continuous and positive on the compact simplex, so a
minimizer $q_*$ of the AUROC gap exists. Let $I_1,\ldots,I_m$ be iid group
indices with probabilities $q_*$, and let
$\widehat q=m^{-1}\sum_{i=1}^m e_{I_i}\in\mathcal G_m$. For every symmetric
matrix $M$,
\[
 m\E[\widehat q^\top M\widehat q]
 =(m-1)q_*^\top Mq_*+\sum_gq_{*,g}M_{gg}.
\]
Write $d=p(q_*)(1-p(q_*))$ and
$T=\sum_gq_{*,g}\pi_g(1-\pi_g)$. Since
$\widehat q^\top B\widehat q\ge\gamma_m\widehat q^\top C\widehat q$,
taking expectations yields
\[
 (m-1)\gamma d+\sum_gq_{*,g}B_{gg}
 \ge\gamma_m\{(m-1)d+T\}.
\]
Moreover $B_{gg}=\pi_g(1-\pi_g)\Delta(e_g)\le b\pi_g(1-\pi_g)$,
so $(m-1)(\gamma-\gamma_m)d\ge(\gamma_m-b)T$. The grid contains all
vertices, hence $\gamma_m\le b$. Concavity of $x(1-x)$ gives $T\le d$;
because the coefficient $\gamma_m-b$ is nonpositive, the preceding inequality
implies $(m-1)(\gamma-\gamma_m)\ge\gamma_m-b$. This is the lower bound.
The upper bound follows from $\mathcal G_m\subseteq\DeltaK$.

For sharp order, use the hierarchy construction with $K\ge m$. A
denominator-$m$ grid point has at most $m$ nonzero coordinates, and its
smallest squared norm is $1/m$, attained by putting mass $1/m$ on $m$
distinct groups. Consequently
\[
 \gamma=\frac12+\frac1{2K}-c,\qquad
 \gamma_m=\frac12+\frac1{2m}-c,\qquad b=1-c.
\]
Their difference is $\tfrac12(m^{-1}-K^{-1})\ge1/(4m)$ when $K\ge2m$.
The bound in Eq.~\meqref{eq:gridbound} equals $1/2-c$, only $1/(2K)$ below
the true minimum. For fixed $m$, its correction is asymptotically attained as
$K$ grows. The probabilistic technique is the classical simplex-grid
argument; the ratio, prevalence cancellation, and attainable comparison
construction are the specialization needed here.
\end{proof}

One immediate corollary clarifies the support hierarchy. If every mixture with
at most $m\ge2$ groups has nonnegative gap, then the grid also has
$\gamma_m\ge0$, while $b\le1$, and hence every mixture has
$\Delta(q)\ge-1/(m-1)$. The theorem does not say that checking only a few
selected sparse mixtures suffices: its premise quantifies over the complete
sparse family. For a desired additive error $\epsilon$, the worst-case bound
$b-\gamma_m\le2$ allows $m\ge1+2/\epsilon$, with a grid size polynomial in
$K$ for fixed $\epsilon$ but potentially impractical at small tolerances.

\subsection{Simultaneous Sampling Statements}
\label{app:statistics}
For distinct groups set
\[
 \widehat B_{gh}=\frac1{n_gn_h}\sum_{i=1}^{n_g}\sum_{j=1}^{n_h}
                         k(Z_{gi},Z_{hj}),
\]
and on the diagonal set
$\widehat B_{gg}=\binom{n_g}{2}^{-1}\sum_{i<j}k(Z_{gi},Z_{gj})$.
At most one of the two label indicators in Eq.~\meqref{eq:kernel} is nonzero,
so $k\in[-1/2,1/2]$, an interval of length one. Both estimators are unbiased.

\begin{proof}[Proof of Theorem~\mref{thm:bands}]
First take $g\ne h$ and suppose $n_g\le n_h$. Independently permute each
sample and pair the $n_g$ observations of the smaller sample with the first
$n_g$ of the larger. The average of the paired kernels has the same mean
$B_{gh}$ and is an average of $m_{gh}=n_g$ independent variables of range
length one. Averaging this statistic over all matchings recovers
$\widehat B_{gh}$: every cross pair has the same inclusion probability.
Hoeffding's exponential-moment bound and Jensen's inequality therefore give
\[
 \E\exp\{\lambda(\widehat B_{gh}-B_{gh})\}
 \le \exp\{\lambda^2/(8m_{gh})\}.
\]
On the diagonal, randomly permute the observations, form
$\lfloor n_g/2\rfloor$ disjoint pairs, and discard the remaining observation
if $n_g$ is odd. The paired terms are independent. Every unordered pair has
equal chance of appearing, so averaging over permutations gives the complete
U-statistic. The same exponential-moment inequality holds with $m_{gg}$.
Chernoff optimization gives two-sided probability at most
$2\exp(-2m_{gh}t^2)$. A union bound over $L$ unique entries across the fixed
comparison family proves the simultaneous entrywise event. Finally,
$q_gq_h\ge0$ for every mixture, so multiplying the entry bounds by these
weights and summing proves Eq.~\meqref{eq:uniform} for the entire simplex
on that single event. Data-dependent selection of $q$ or $\Q$ cannot leave
this event. It cannot enlarge the prespecified score or group family for
which the event was proved.
\end{proof}

\paragraph{Paired variance-sensitive intersection.}
We use a known empirical-Bernstein inequality \citep[Theorem~4]{maurer2009empirical}:
for $m\ge2$ iid variables in an interval of length one, sample mean
$\bar W$ and unbiased sample variance $s_W^2$, a two-sided interval of error
probability at most $\alpha$ has radius
\[
 r_{\rm EB}=\sqrt{\frac{2s_W^2\log(4/\alpha)}m}
       +\frac{7\log(4/\alpha)}{3(m-1)}.
\]
It follows by applying the cited one-sided inequality to $W$ and to $-W$,
each with failure probability $\alpha/2$. Use one matching per matrix entry,
fixed independently of observed labels and scores, and the kernel values on
those disjoint pairs. They are iid under the assumptions of
Theorem~\mref{thm:bands}. Allocate total error $\delta/2$ to all-pairs
Hoeffding intervals, replacing their logarithm by $\log(4L/\delta)$; allocate
the remaining $\delta/2$ to the $L$ matched-pair intervals, using
$\log(8L/\delta)$ above. Intersect the two intervals with $[-1/2,1/2]$.
The union bound proves simultaneous coverage at least $1-\delta$. The
intersection need not be centered at the complete U-statistic. If numerical
or off-event computations produce an empty intersection, the implementation
returns $[-1/2,1/2]$ and records a diagnostic, a conservative fail-open rule.
The inequalities are standard; their paired construction here keeps the
correlation between the two competing methods.

\paragraph{Exact arithmetic for statistical endpoints.}
Kernel values are multiples of $1/4$. The matching means and variances are
computed from integer sums as rational numbers. The code encloses logarithms
and square roots using 60-digit decimal arithmetic, correctly rounded
transcendental operations followed by an outward step, and upward rounding
for positive arithmetic. Intersections are then rational and can be passed
directly to the exact optimizer. Floating outputs are rounded outward for
display. Statistical validity still depends on the sampling assumptions;
arithmetic certification cannot supply missing independence.

\paragraph{Quantitative gaps with uncertain prevalence.}
An independent error allocation $\delta_\pi$ gives simultaneous bounds
\[
 \underline\pi_g=\max\left\{0,\widehat\pi_g-
     \sqrt{\frac{\log(2K/\delta_\pi)}{2n_g}}\right\},\qquad
 \overline\pi_g=\min\left\{1,\widehat\pi_g+
     \sqrt{\frac{\log(2K/\delta_\pi)}{2n_g}}\right\}.
\]
For every $q$, let $p_L=q^\top\underline\pi$, $p_U=q^\top\overline\pi$,
$d_L=\min\{p_L(1-p_L),p_U(1-p_U)\}$, and
$d_U=\max_{p\in[p_L,p_U]}p(1-p)$. The latter equals $1/4$ if the interval
contains $1/2$, and otherwise its larger endpoint value. If $d_L>0$, take
the smallest and largest of the four ratios
$N/d$ with $N\in\{q^\top\underline Bq,q^\top\overline Bq\}$ and
$d\in\{d_L,d_U\}$, then intersect with $[-1,1]$. All these bounds hold
simultaneously for all mixtures with failure probability at most
$\delta+\delta_\pi$. If $d_L=0$, return $[-1,1]$; do not divide by an
estimated prevalence. The sign certificates need no prevalence interval.
The delivered prevalence and ratio convenience functions use floating-point
arithmetic and are not presented as rational certificates. The exact
statistical certification path uses the rational sign-envelope endpoints;
the completed quantitative finite-bank calculations use exact prevalences.

\paragraph{Independent clusters.}
Suppose iid independent units $W_i$ each contain group records $Z_{ig}$,
which may be arbitrarily dependent within unit. Use the symmetric kernel
\[
 K_{gh}(W_i,W_j)=\tfrac12\{k(Z_{ig},Z_{jh})+k(Z_{jg},Z_{ih})\},
 \qquad i\ne j.
\]
Its expectation is $B_{gh}$ and its range has length one. A complete
degree-two U-statistic over independent units therefore satisfies the same
proof with $m=\lfloor n/2\rfloor$ for every entry. Within-unit cross pairs
must be excluded. If units contain varying numbers of questions, one must
also specify how units and questions are sampled and weighted; merely
counting all questions as independent does not instantiate this theorem.
The real score-bank analysis consequently makes finite-bank statements,
keeps story and duplicate-input clusters together in its split, and does
not report an iid population guarantee for its rows.

\subsection{Marginal Alignment and Its Approximation Error}
\label{app:alignment}
For any score with law $F$, define
$F^{\rm mid}(t)=\Prob(S<t)+\tfrac12\Prob(S=t)$. For iid $S,S'$, the identity
$\phi(S,S')+\phi(S',S)=1$ gives $\E[F^{\rm mid}(S)]=1/2$ without continuity.
For independent, not necessarily identically distributed $S_g,S_h$ it gives
\begin{equation}
 \E_g F_h^{\rm mid}(S_g)+\E_h F_g^{\rm mid}(S_h)=1.\label{eq:crossmid}
\end{equation}

\begin{lemma}[Placement representation]
\label{lem:placement}
For a binary label with prevalence $p\in(0,1)$ and marginal score law $F$,
\[
 p(1-p)\AUC(S;Y)=\E[Y F^{\rm mid}(S)]-p^2/2.
\]
\end{lemma}
\begin{proof}
Let $(Y',S')$ be an independent copy. The first expectation is
$\E[Y\phi(S,S')]$. Splitting on $Y'$ yields the desired positive-negative
term plus $\E[YY'\phi(S,S')]$. Conditional on $Y=Y'=1$, the two positive
scores are iid, so their tie-adjusted ordering probability is $1/2$. The
second term is therefore $p^2/2$.
\end{proof}

\begin{proof}[Proof of Theorem~\mref{thm:alignment}]
Let $F_{s,q}^{\rm mid}=\sum_gq_gF_{s,g}^{\rm mid}$. Applying
Lemma~\ref{lem:placement} to the two methods and subtracting gives
\[
 N(q):=p(q)(1-p(q))\Delta(q)
 =\sum_gq_g\E_g\!\left[Y\{F_{A,q}^{\rm mid}(S_A)
                             -F_{B,q}^{\rm mid}(S_B)\}\right].
\]
Applying the same lemma within group $g$ gives
$a_g=\E_g[Y\{F_{A,g}^{\rm mid}(S_A)-F_{B,g}^{\rm mid}(S_B)\}]$.
When each method's marginal is common across groups, subtracting these
identities immediately gives $N(q)=q^\top a$. Pure-group necessity and the
nonnegative linear combination establish the dominance equivalence.

For the error bound, for each method and unordered group pair define
\[
 R_{s,gh}=\E_g[Y\{F_{s,h}^{\rm mid}(S_s)-F_{s,g}^{\rm mid}(S_s)\}]
          +\E_h[Y\{F_{s,g}^{\rm mid}(S_s)-F_{s,h}^{\rm mid}(S_s)\}].
\]
Expanding the mixture CDF and pairing the directed terms yields
$N(q)-q^\top a=\sum_{g<h}q_gq_h(R_{A,gh}-R_{B,gh})$.
The absolute value of $R_{s,gh}$ is at most
$(\pi_g+\pi_h)\delta_{s,gh}$. By Eq.~\eqref{eq:crossmid}, replacing $Y$
by the constant one in its definition gives zero. Replacing $Y$ by
$-(1-Y)$ therefore gives exactly the same $R_{s,gh}$, and supplies the
second bound $(2-\pi_g-\pi_h)\delta_{s,gh}$. Take the smaller bound and
apply the triangle inequality over methods and group pairs. This proves
Eq.~\meqref{eq:alignmenterror}, including arbitrary ties.
\end{proof}

\paragraph{An entrywise envelope.}
Define $H_{gh}=(a_g+a_h)/2$,
$E_{gg}^{\rm align}=0$, and, for $g\ne h$,
\[
 E_{gh}^{\rm align}=\tfrac12 w_{gh}
          (\delta_{A,gh}+\delta_{B,gh}).
\]
The coefficient identity in the proof gives
$B_{gh}-H_{gh}=\tfrac12(R_{A,gh}-R_{B,gh})$, hence
$H-E^{\rm align}\le B\le H+E^{\rm align}$ entrywise. The same optimizers
can use this envelope. It requires valid within-group and marginal-law
information; plug-in empirical marginals alone are not a population envelope.

Let $\delta_s=\max_{g<h}\delta_{s,gh}$. Summing the positive-class bound
instead of taking pairwise minima gives
$|N(q)-q^\top a|\le(\delta_A+\delta_B)p(q)$; the negative-class version
gives the same expression with $1-p(q)$. Dividing by the denominator yields
\begin{equation}
 \left|\Delta(q)-\frac{q^\top a}{p(q)(1-p(q))}\right|
 \le\frac{\delta_A+\delta_B}{\max\{p(q),1-p(q)\}}
 \le2(\delta_A+\delta_B).\label{eq:alignmentglobal}
\end{equation}
This avoids an artificial explosion at rare class prevalences. The sharper
pairwise bound vanishes at pure groups and can be substantially smaller.

\paragraph{Oracle rank alignment and ties.}
For method $s$ in group $g$, let $V$ be a fresh independent uniform variable,
independent of the label, and transform
\[
 U_s=F_{s,g}(S_s^-)+V\{F_{s,g}(S_s)-F_{s,g}(S_s^-)\}.
\]
The distributional probability-integral transform makes $U_s$ uniform.
For two unequal score values their transformed intervals are ordered. When
the original scores tie, the independent uniform positions rank each way
with probability $1/2$. Thus the original tie-adjusted within-group AUROC
is preserved in expectation over the added randomization. This is an oracle
construction, not a claim that an estimated empirical step transform has
the same property on new data. Finite step transforms can create additional
ties; the completed control reports those changes and separately tests a
strict monotone interpolation.

\paragraph{An unlabeled finite-calibration implication.}
For each fixed method and group, suppose an independent calibration sample
of size $n_g^{\rm cal}$ estimates its marginal CDF. The classical DKW--Massart
bound \citep{massart1990tight} and a union bound over $M$ methods and $K$
groups give, with probability at least $1-\alpha$,
\[
 \|\widehat F_{s,g}-F_{s,g}\|_\infty\le
 \epsilon_g=\sqrt{\frac{\log(2MK/\alpha)}{2n_g^{\rm cal}}}
 \quad\text{for all }s,g.
\]
The same inequality holds for left limits. Coupling the estimated randomized
transform with the oracle transform using the same $V$ gives
$|\widehat U_s-U_s|\le\epsilon_g$. Since the oracle is uniform, the CDF
of $\widehat U_s$ is within $\epsilon_g$ of the uniform CDF on $[0,1]$.
Its midpoint CDF has the same bound, being the average of the right and
left limits. Therefore for the transformed scores,
$\delta_{A,gh}+\delta_{B,gh}\le2(\epsilon_g+\epsilon_h)$.
Substitution into the alignment envelope gives a valid calibration-error
bound. The within-group AUROCs in that envelope are those of the
\emph{transformed} scores; finite empirical calibration need not preserve
the original AUROCs. Obtaining their population confidence bounds still
requires labeled evaluation data independent of the fitted transforms.

\paragraph{Negative-class calibration as a related boundary.}
If, for each method, its negative-class score law is invariant across groups,
then $d_{gh}$ is independent of $h$ and equals $\Delta(e_g)$. Thus
$\Delta(q)=\sum_g[q_g\pi_g/p(q)]\Delta(e_g)$. This is the familiar
negative-placement/covariate-adjusted ROC setting
\citep{janes2009adjusting}; cf.\ the within/cross-group AUC decomposition of
\citet{matos2026collapsibility}. The marginal-alignment
identity in Eq.~\meqref{eq:aligned} has different weights and does not require
negative-class labels to define the population transform. Neither boundary
makes preservation of useful cross-group risk information automatic.

%% file: sec/supp_computation.tex
\section{Algorithms, Optimization Proofs, and Software}
\label{app:computation}
\subsection{Exact Input Representation}
For method $s$ and groups $g,h$, let
\[
 W^s_{gh}=\sum_{i:Y_{gi}=1}\sum_{j:Y_{hj}=0}
   \left(2\ind{S_{sgi}>S_{shj}}+\ind{S_{sgi}=S_{shj}}\right).
\]
This is an integer and can be computed by sorting negative scores and using
binary searches for the two tie boundaries. With $n_g$ total rows, the exact
finite-bank matrix entries are
\begin{equation}
 B^{\rm emp}_{gh}=
 \frac{(W^A_{gh}-W^B_{gh})+(W^A_{hg}-W^B_{hg})}{4n_gn_h}.
\end{equation}
When $g=h$, the two terms in the numerator are identical. For unbiased
population estimation only, replace the diagonal denominator by
$4n_g(n_g-1)$. Empirical prevalences are the exact fractions $n_g^+/n_g$.
These formulas prevent floating score values from introducing numerical
ambiguity into the comparison probabilities: score ordering is frozen by the
released values, while probabilities and optimization use exact integers and
fractions. Ties are determined from the stored values without arbitrary
rounding or a tolerance.

\subsection{Global Quadratic Brackets}
Let $D$ be symmetric and rational and let the nonempty rational mixture
polytope be $\Q=\{q\ge0:\one^\top q=1,Hq\le h\}$. Every cell is the
intersection of $\Q$ with a rational coordinate box. Let its vertices be
$V=\{v_1,\ldots,v_J\}$. Define
$L(\mathcal P)=\min_{i,j}v_i^\top Dv_j$ and
$U(\mathcal P)=\min_i v_i^\top Dv_i$.

\begin{proposition}[Cell bounds and convergence]
\label{prop:cells}
For every nonempty cell,
\[
 L(\mathcal P)\le\min_{q\in\mathcal P}q^\top Dq\le U(\mathcal P),
 \qquad
 U(\mathcal P)-L(\mathcal P)
 \le\tfrac12\|D\|_2\operatorname{diam}(\mathcal P)^2.
\]
Maintaining a covering partition, pruning cells only when their lower bound
is at least a feasible incumbent, and bisecting the longest coordinate
interval yields valid global brackets at every completed iteration. It
reaches any positive requested absolute bracket width after finitely many
refinements if not interrupted.
\end{proposition}
\begin{proof}
Every feasible $q$ is a convex combination $\sum_i\alpha_i v_i$, hence
$q^\top Dq=\sum_{i,j}\alpha_i\alpha_jv_i^\top Dv_j$ is a convex combination
of the bilinear vertex values. This proves the lower bound, and feasible
vertices prove the upper bound. For any vertices $u,v$,
\[
 u^\top Dv=\tfrac12\{u^\top Du+v^\top Dv-(u-v)^\top D(u-v)\}
 \ge U(\mathcal P)-\tfrac12\|D\|_2\|u-v\|_2^2.
\]
Taking the minimum gives the gap bound. The minimum active-cell lower bound
is a global lower bound because the active cells together with correctly
pruned cells cover $\Q$ and every pruned objective is at least the feasible
incumbent. The incumbent itself is an upper bound. Longest-coordinate
bisection shrinks cell diameters uniformly with depth: after sufficiently
many levels every coordinate width is below a prescribed positive number.
The gap bound then falls below the requested tolerance. There are finitely
many nodes before that depth. Best-bound refinement cannot leave a
larger-gap relevant cell forever while processing only deeper cells, since
cells of sufficiently small gap either close the global bracket or are
pruned by the current incumbent. Thus finite-tolerance termination follows.
Exact zero-margin decisions are not implied by this argument.
\end{proof}

The code eliminates $q_K$ and enumerates all sets of $K-1$ linearly
independent active inequalities to obtain exact vertices, including
lower-dimensional feasible sets. It solves each rational system and checks
every original inequality exactly. Cached inverse bases reduce repeated
work during bisection. This classical partition method is described to make
its certificates auditable, not claimed as a new global-optimization method.

\paragraph{Reference audit algorithm.}
\begin{enumerate}[leftmargin=2em]
\item Freeze score methods, target labels, groups, reference mixture and allowed
mixture domain. Compute all integer rank counts and rational matrices.
\item For a desired advantage $\tau$, form $D=B-\tau C$. For a population
sign statement substitute the appropriate simultaneous lower or upper
matrix; do not substitute an empirical prevalence into a population margin.
\item Initialize the feasible incumbent at a polytope vertex and insert the
initial cell with its bilinear lower bound into a priority queue.
\item Refine the cell with the smallest lower bound by exact midpoint
bisection. Discard empty children; update the feasible incumbent and prune
children whose lower bound cannot improve it.
\item Return nonnegative dominance only if the global lower bound is
nonnegative. Return a reversal only with a feasible negative witness.
At a budget stop, return the valid interval and \emph{unresolved} if needed.
\item For small $K$, the stationary-face procedure below can settle exact
zero cases. For a quantitative gap or positive-margin radius, use the
certified quadratic oracle inside the stated bisection transformations.
\end{enumerate}

\subsection{Exact Stationary-Face Enumeration}
After eliminating the simplex equality, write
$f(z)=z^\top Rz+2r^\top z+c$ on the compact polytope $P=\{z:Az\le b\}$.
For every independent subset of active equalities $Ez=e$, choose a basis $N$
of $\ker E$ and solve
\begin{equation}
 Ez=e,\qquad N^\top(Rz+r)=0.\label{eq:stationary}
\end{equation}
Intersect this affine solution set with $P$, and retain one rational feasible
point if the intersection is nonempty. Include all vertices. Minimize $f$
over the retained points.

\begin{proposition}[Exact fallback]
This finite enumeration obtains the exact global minimum, including singular
Hessians and lower-dimensional polytopes.
\end{proposition}
\begin{proof}
Let $z_*$ be a global minimizer, and let $F$ be its minimal containing face.
Then $z_*$ is in the relative interior of $F$. An independent basis of its
active equalities is among the enumerated sets. Both signs of a small
tangential displacement are feasible, so first-order optimality gives
Eq.~\eqref{eq:stationary}. If two solutions $z,z'$ share this stationary
affine set, put $d=z'-z$. Then $Ed=0$, $N^\top Rd=0$, and
$N^\top(Rz+r)=0$. Since $d$ lies in the span of $N$,
$d^\top Rd=d^\top(Rz+r)=0$, and therefore $f(z')=f(z)$. Any feasible
point of this stationary set has the global-minimum value. The compact
intersection of rational affine equations and $P$, when nonempty, has a
rational vertex, obtainable by adding enough independent active inequalities.
The code finds such a point. Every retained point is feasible, so no retained
value can be below the global minimum, while the minimizing face supplies
equality. No assertion that indefinite quadratic minima occur only at
vertices is used.
\end{proof}

\subsection{Normalized Gaps, Radii, and Restricted Grids}
For any trial $\tau$, positivity of $q^\top Cq$ gives
\[
 \min_{q\in\Q}\Delta(q)\ge\tau
 \quad\Longleftrightarrow\quad
 \min_{q\in\Q}q^\top(B-\tau C)q\ge0.
\]
The normalized-gap routine maintains a dyadic bracket in $[-1,1]$. A
nonnegative quadratic minimum raises the lower bound; a negative feasible
witness lowers the upper bound. If an interrupted oracle cannot decide the
sign, the outer routine retains its current bracket. It never infers the
worst normalized gap by minimizing the numerator alone.

For equal-mass probability vectors,
\[
 \TV(q,q_0)=\max_{T\subseteq\{1,\ldots,K\}}\sum_{g\in T}(q_g-q_{0,g}).
\]
Indeed the maximizing set is the positive-coordinate-difference set, and its
sum is half the $\ell_1$ norm. Thus a TV ball is a rational polytope with
$2^K-2$ nontrivial subset inequalities. At a trial radius $r$, minimize
$q^\top(B+\kappa C)q$ over its intersection with $\Q$, for $\kappa>0$.
A strictly positive lower bound excludes a reversal throughout that ball;
a nonpositive feasible value exhibits a reversal of at least $\kappa$.
The feasible witness's exact TV distance supplies the upper bracket endpoint.
The lower endpoint is valid by nesting of TV balls. If no reversal exists
over all $\Q$, the radius is infinite. An isolated exact tie cannot satisfy
this positive-margin target. The implementation rejects a zero-margin
radius request rather than silently interpreting a tie as a strict reversal.

The corrected-grid theorem extends to a nonempty rational polytope whose
vertices are $v_1,\ldots,v_J$. Treat $P_{v_j}=\sum_gv_{j,g}P_g$ as the
component distributions, with prevalences $\pi^\top v_j$. A mixture over
these components has the same AUROC as the corresponding point of $\Q$.
Applying Theorem~\mref{thm:grid} to the equally weighted averages of $m$
vertices gives its bound with $b=\max_j\Delta(v_j)$. An arbitrary coordinate
grid intersected with a polytope is not this grid and need not satisfy the
same correction. For $q_g\ge0.05$ with four groups, the transformation
$q=0.05\one+0.8t$, $t\in\Delta_4$, is exactly a vertex-mixture grid.

Our sharp-order statement is dimension-uniform over varying $K$, not a
claim that every fixed instance converges at rate only $1/m$. Faster
instance-dependent quadratic grid bounds are known
\citep{deklerk2015grid}. The hierarchy used for our lower bound has
$K\ge2m$ and is fully compatible with those results.

\subsection{Comparable Baselines}
The audit compares the same two frozen scores and the same group law under
each baseline. Within-group signs use only the diagonal. Entrywise cross-AUC
nonnegativity certifies the larger class of independently shiftable positive
and negative mixtures. Exhaustive two-group tests optimize all simplex
edges. The uncorrected denominator-40 grid supplies feasible witnesses;
the corrected grid supplies a valid lower bound. Exact quadratic certificates
settle the small-$K$ sign problem. The alignment envelope uses only
within-group gaps and pairwise marginal-CDF discrepancies.

For a distributional-drift comparator, we implement the paired-loss
Hellinger certificate of \citet{weber2022certifying}, adapted to the same
positive-negative pair law. At $q_0$, put
$D=\phi(S_A^+,S_A^-)-\phi(S_B^+,S_B^-)\in[-1,1]$ and
$L=(D+1)/2\in[0,1]$, with mean $\mu$ and variance $v$. Define the
class-conditional mixture weights $r^\pm(q)$ as in Sec.~\mref{sec:formulation} of the main paper.
The labeled group-pair Hellinger affinity is
\[
 a(q)=\left(\sum_g\sqrt{r_g^+(q_0)r_g^+(q)}\right)
       \left(\sum_h\sqrt{r_h^-(q_0)r_h^-(q)}\right).
\]
It is exact with group identities retained and a lower bound on affinity
after dropping them. The Gramian bound gives the valid paired gap lower bound
\begin{equation}
 \Delta(q)\ge
 2\left[\max\left\{0,a(q)\sqrt\mu-
              \sqrt{1-a(q)^2}\sqrt{v/\mu}\right\}\right]^2-1,
 \label{eq:hellingerbaseline}
\end{equation}
with the conservative value $-1$ when $\mu=0$. The positive part must be
taken before squaring; otherwise the bound can be invalid at large shifts.
The implementation computes the exact five-point empirical distribution of
$D$ in blocks, retaining ties and method pairing. For a whole TV region it
also exposes the analytic affinity lower bound
$a_r=(1-\min\{1,r/p(q_0)\})(1-\min\{1,r/(1-p(q_0))\})$.
This region conversion is explicitly conservative; we do not claim it is an
optimal implementation of Hellinger-ball certification. Pointwise comparisons
use the exact group-pair affinity. Empirical moments here describe the finite
bank; population use requires simultaneous moment and prevalence uncertainty.

\paragraph{Reason for Eq.~\eqref{eq:hellingerbaseline}.}
Let $\nu=P+Q$, $\psi_P=\sqrt{dP/d\nu}$ and
$\psi_Q=\sqrt{dQ/d\nu}$. These are unit vectors in $L^2(\nu)$ with
inner product $a$. If $a<1$, write
$\psi_Q=a\psi_P+\sqrt{1-a^2}\,w$, where $w$ is a unit vector
orthogonal to $\psi_P$. Then
\[
 \int L\psi_P\psi_Q\,d\nu
 =a\mu+\sqrt{1-a^2}\int(L-\mu)\psi_Pw\,d\nu
 \ge a\mu-\sqrt{(1-a^2)v}.
\]
Cauchy--Schwarz also bounds this integral above by
$\sqrt{\E_P L\,\E_Q L}$. For $\mu>0$, taking the positive part,
dividing by $\sqrt\mu$, and squaring proves the bound. If $a=1$, the
two laws coincide and the formula is equality; if $\mu=0$, the stated
lower bound is trivial. This proof includes mutually singular components.
It is the standard Hellinger/Gramian argument, used as a comparison baseline
rather than claimed as a contribution.

\subsection{Reproducibility}
The implementation comprises the exact optimizer, the U-statistic confidence
routines, the alignment and Hellinger comparators, a generic audit interface
for score tables, the prespecified protocols, integer rank counts, complete
pair tables, and the figure code; it will be released with the paper. The
public score-bank revision and checksums of all inputs are recorded. No
training or generation was rerun for the completed language-model study.
The tests cover tied ranks, unequal prevalences, agreement of weighted
AUROCs with an independent library, interior and singular minima,
non-dyadic zero minima, constraint feasibility, and the analytic hierarchy
radius.

%% file: sec/supp_empirical_details.tex
\section{Completed Score-Bank Audit: Protocol, Evidence, and Counterevidence}
\label{app:empirical-details}

\subsection{Source, sampling units, and frozen analysis}

We analyze author-released LM-Polygraph score banks, associated with the
benchmark of \citet{vashurin2025polygraph}. The source is the public
\texttt{LM-Polygraph/managers} dataset at Hugging Face, at a pinned
revision recorded with the code to be released: the four Stable LM 2 12B result
files for MMLU, CoQA, GSM8K, and TriviaQA. The source paper identifies the white-box model
as Stable LM 2 12B \emph{base}, without instruction tuning. We use the released
generations and scores; we did not run new model inference. The banks do not
record the original inference code or checkpoint revision, and the release
contains no second base-model bank.

Each downloaded file is pinned by its SHA-256 checksum, size, and source
revision, and is read with a restricted data-only loader. The public
repository metadata lists an MIT license. The source benchmark and constituent datasets retain their
respective attribution requirements.

All 35 common uncertainty methods are retained (Table~\ref{tab:method-inventory}).
All 11,019 rows have finite values for all these scores and both released
endpoint arrays; the common-complete-case policy excludes zero rows. Scores
retain their published orientation, with larger values indicating greater
uncertainty. No orientation is selected using AUROC. Every released row,
including repeated inputs, remains in the empirical target population.

\begin{table}[ht]
\centering\small
\caption{Audit sample sizes. Discovery/confirmation assignment is at the
input/story-cluster level, so row counts need not be equal. Error counts refer
to the released Accuracy endpoint.}
\label{tab:bank-counts}
\begin{tabular}{lrrrrrrr}
\toprule
Task & Rows & Discovery & Confirmation & Disc. clusters & Conf. clusters & Disc. errors & Conf. errors\\
\midrule
MMLU & 5700 & 2840 & 2860 & 2830 & 2856 & 1287 & 1305\\
CoQA & 2000 & 1059 & 941 & 260 & 226 & 651 & 608\\
GSM8K & 1319 & 671 & 648 & 671 & 648 & 300 & 290\\
TriviaQA & 2000 & 1035 & 965 & 978 & 912 & 448 & 415\\
\midrule
Total & 11019 & 5605 & 5414 & 4739 & 4642 & 2686 & 2618\\
\bottomrule
\end{tabular}
\end{table}

We assign exact duplicate inputs to one cluster. For CoQA, we instead extract
the complete shared story between the unique \texttt{Story:} marker and its
first following \texttt{Question:} marker; all questions from a story remain
together. Parsing is checked for every row. After normalizing line endings and
stripping outer whitespace, SHA-256 hashes of a fixed protocol identifier,
task name, and cluster text assign clusters to discovery or confirmation
according to whether the first digest byte is below 128. This produces the
counts in Table~\ref{tab:bank-counts}. It does not establish independence of
benchmark questions beyond these observed clusters.

The primary protocol was fixed after inspecting schemas, finite counts,
duplicate inputs, and example prompt formats, but before computing new score
comparisons. It is an outcome-blind retrospective protocol, not prospective
preregistration. The reference mixture is $q_0=(1/4,1/4,1/4,1/4)$, in task order
MMLU, CoQA, GSM8K, TriviaQA. For each of the 595 unordered pairs, $A$ is the
discovery reference winner. An exhaustive grid of integer compositions of 40
into four parts selects a discovery stress mixture; the same mixture is then
evaluated on confirmation. The secondary domain is $q_g\geq0.05$, using
$q=0.05\mathbf1+0.8t$ for grid point $t$. All pairs and null outcomes are
reported. Exact rational optimization independently audits both continuous
domains, using integer twice-win counts and the empirical-product matrix.
Pairs share methods and observations; their counts are descriptive, not 595
independent replications or a random sample from a population of UQ methods.

\begin{table}[ht]
\centering\scriptsize
\caption{Complete method inventory, with stable identifiers used below. These
are the exact released estimator names; all 35 are analyzed without selection.}
\label{tab:method-inventory}
\begin{tabular}{rlrl}
\toprule
ID & Estimator & ID & Estimator\\
\midrule
1 & \texttt{MaximumSequenceProbability} & 19 & \texttt{DegMat\_NLI\_score\_entail}\\
2 & \texttt{Perplexity} & 20 & \texttt{DegMat\_NLI\_score\_contra}\\
3 & \texttt{MeanTokenEntropy} & 21 & \texttt{DegMat\_Jaccard\_score}\\
4 & \texttt{MeanPointwiseMutualInformation} & 22 & \texttt{Eccentricity\_NLI\_score\_entail}\\
5 & \texttt{MeanConditionalPointwiseMutualInformation} & 23 & \texttt{Eccentricity\_NLI\_score\_contra}\\
6 & \texttt{CCP} & 24 & \texttt{Eccentricity\_Jaccard\_score}\\
7 & \texttt{PTrue} & 25 & \texttt{SemanticEntropy}\\
8 & \texttt{PTrueSampling} & 26 & \texttt{SAR}\\
9 & \texttt{MonteCarloSequenceEntropy} & 27 & \texttt{TokenSAR}\\
10 & \texttt{MonteCarloNormalizedSequenceEntropy} & 28 & \texttt{SentenceSAR}\\
11 & \texttt{LexicalSimilarity\_rouge1} & 29 & \texttt{RenyiNeg}\\
12 & \texttt{LexicalSimilarity\_rouge2} & 30 & \texttt{FisherRao}\\
13 & \texttt{LexicalSimilarity\_rougeL} & 31 & \texttt{MahalanobisDistanceSeq\_decoder}\\
14 & \texttt{LexicalSimilarity\_BLEU} & 32 & \texttt{RelativeMahalanobisDistanceSeq\_decoder}\\
15 & \texttt{NumSemSets} & 33 & \texttt{RDESeq\_decoder}\\
16 & \texttt{EigValLaplacian\_NLI\_score\_entail} & 34 & \texttt{PPLMDSeq\_decoder}\\
17 & \texttt{EigValLaplacian\_NLI\_score\_contra} & 35 & \texttt{PPLRMDSeq\_decoder}\\
18 & \texttt{EigValLaplacian\_Jaccard\_score} & & \\
\bottomrule
\end{tabular}
\end{table}

\subsection{Endpoint semantics and label sensitivity}

The primary binary endpoint is $Y=1-\texttt{Accuracy}$ from the release.
It is task-dependent textual agreement, not uniformly human semantic
correctness. We independently reproduce every primary label from the stored
generations and targets (Table~\ref{tab:label-semantics}). These checks establish
the released labels' observable semantics; they do not recover the unknown
historical execution configuration.

\begin{table}[ht]
\centering\small
\caption{Exact reproduction of released Accuracy and changes under the frozen
task-parser sensitivity. Each primary reconstruction has zero disagreements.}
\label{tab:label-semantics}
\begin{tabular}{lp{9.8cm}r}
\toprule
Task & Reproducing rule for released Accuracy & Parser changes\\
\midrule
MMLU & Case-sensitive equality after stripping output and target letter. & 7/5700\\
CoQA & Case-sensitive equality after stripping output and the single stored answer; no punctuation, article, or case normalization. & 676/2000\\
GSM8K & Greedily remove target prefix through \texttt{\#\#\#\# } and output prefix through \texttt{The answer is }; strip, lowercase, remove ASCII punctuation, and compare. & 7/1319\\
TriviaQA & Strip, lowercase, remove ASCII punctuation, and compare with any provided target alias. & 28/2000\\
\bottomrule
\end{tabular}
\end{table}

The first secondary endpoint uses $Y=\ind{\texttt{AlignScore}<0.5}$. The second
uses task-specific parsers fixed before comparisons: a leading MMLU answer
letter; exact Decimal comparison of GSM8K's final extracted number, preferring
text following the final \texttt{\#\#\#\#} marker; official TriviaQA normalization
and alias matching; and normalized single-reference CoQA token-F1 at least
0.5. Missing parses count as incorrect. CoQA's latter rule is a sensitivity,
not its official multireference evaluation. AlignScore and the parser labels
are also proxies. The substantial CoQA change is therefore scientifically
material, rather than an implementation nuisance.

\begin{table}[ht]
\centering\small
\caption{Exact continuum classification. Each entry counts mixture-dependent
pairs out of 595; the remaining pairs have a universal ordering. ``Interior''
means every task has at least 0.05 mass. Each endpoint uses its own
discovery-reference orientation, without changing score direction.}
\label{tab:all-endpoint-certificates}
\begin{tabular}{lrrrrrr}
\toprule
 & \multicolumn{3}{c}{Full simplex} & \multicolumn{3}{c}{Interior}\\
Endpoint & Discovery & Confirmation & Full bank & Discovery & Confirmation & Full bank\\
\midrule
Released Accuracy & 419 & 422 & 419 & 350 & 358 & 355\\
AlignScore threshold & 415 & 406 & 404 & 294 & 293 & 288\\
Task-parser sensitivity & 388 & 396 & 387 & 287 & 297 & 292\\
\bottomrule
\end{tabular}
\end{table}

Table~\ref{tab:all-endpoint-certificates} reports the 10,710 exact certificates
across endpoints, splits, and domains. For the primary discovery bank, 176
pairs exhibit universal dominance and 419 exhibit mixture-dependent ranking.
The grid detects every primary discovery sign reversal also established by
continuous optimization. Grid gaps remain feasible witness values, not exact
minima of the normalized gap.

Of the 419 primary discovery reversals, 390 remain negative at the unchanged
witness on confirmation. Among 238 pairs whose discovery winner also wins
within every task, 62 reverse, and 55 retain a negative gap at that witness on
confirmation. Only 40 of those 62 also retain all four within-task wins on
confirmation. These are different properties and should not be conflated.

The frozen illustration rule chooses the five largest discovery reversals
among pairs with reference advantage at least 0.005 and a strictly positive
gap in every discovery task. No confirmation outcome enters this rule.
Table~\ref{tab:fixed-examples} retains all five, including unfavorable label
sensitivities. For example, the largest raw reversal disappears under both
secondary endpoints, whereas the second and fourth remain negative.

\begin{table}[ht]
\centering\small
\caption{Confirmation gaps for all five frozen illustrations, at the
\emph{same primary discovery witness}. IDs refer to
Table~\ref{tab:method-inventory}; task order is MMLU/CoQA/GSM8K/TriviaQA.
Intervals are descriptive pointwise 95\% cluster-bootstrap intervals for the
released-Accuracy gap, not simultaneous finite-sample guarantees.}
\label{tab:fixed-examples}
\begin{tabular}{lclrr}
\toprule
$A$ versus $B$ & $q$ & Accuracy gap [interval] & AlignScore gap & Parser gap\\
\midrule
3 vs 31 & $(.525,.475,0,0)$ & $-.1234\;[-.1627,-.0861]$ & $+.3067$ & $+.2163$\\
1 vs 20 & $(0,0,.5,.5)$ & $-.0793\;[-.1110,-.0473]$ & $-.0475$ & $-.0759$\\
13 vs 10 & $(0,.525,.475,0)$ & $-.1017\;[-.1285,-.0742]$ & $+.1439$ & $+.0956$\\
6 vs 23 & $(0,0,.5,.5)$ & $-.0609\;[-.0872,-.0357]$ & $-.0497$ & $-.0554$\\
22 vs 18 & $(0,.475,.525,0)$ & $-.1148\;[-.1418,-.0876]$ & $+.0735$ & $+.0092$\\
\bottomrule
\end{tabular}
\end{table}

The same sensitivity is computed for every primary pair, with no
reorientation or witness reselection. Among the 419 primary discovery
reversals, confirmation remains negative for 390 under Accuracy, 206 under
AlignScore, 251 under the parsers, and 188 under all three. Among the 62
primary all-task-winner reversals, the respective counts are 55, 20, 13, and
13. Requiring confirmation within-task dominance as well gives 40, 0, and 9
for the individual endpoints. These fixed-witness results are more
conservative than reoptimizing separately for each endpoint.

\subsection{Posthoc calibration and strong stress baselines}

After the primary outcomes, but before computing calibration outcomes, we
fixed four controls. Each method is transformed within each task using
discovery data only: the marginal empirical mid-CDF, the correct-example
mid-CDF, and strictly monotone interpolation/extrapolation versions of both.
The mid-CDF uses $[\#(s<x)+\#(s=x)/2]/n$. The strict versions interpolate
between unique calibration values, extrapolate with the end slopes, and use
an affine fallback for a constant calibration sample. Monotone floating-point
tie correction preserves equal original scores while separating distinct
ones. No label-dependent performance tuning is performed. The strict
versions preserve all 140 within-task AUROCs exactly; mid-CDF ties change them
by at most 0.000236 (marginal) and 0.000535 (correct-example).

\begin{table}[ht]
\centering\small
\caption{Posthoc calibration on the \emph{same confirmation bank}. The
within-task-winning subset has 239 pairs for each row. The strict transforms
preserve every within-task ranking, so the reduction from 66 reversals to zero
cannot be explained by changing within-task discrimination.}
\label{tab:calibration-controls}
\begin{tabular}{lrrrr}
\toprule
Scores & Mixture-dependent & Interior & All-task winners & Winners that reverse\\
\midrule
Raw & 422/595 & 358/595 & 239 & 66\\
Marginal mid-CDF & 347/595 & 268/595 & 239 & 0\\
Marginal strict rank & 347/595 & 268/595 & 239 & 0\\
Correct-example mid-CDF & 348/595 & 264/595 & 239 & 1\\
Correct-example strict rank & 348/595 & 264/595 & 239 & 1\\
\bottomrule
\end{tabular}
\end{table}

This is strong limiting evidence: groupwise score alignment removes every
all-task-winner reversal under marginal calibration
(Table~\ref{tab:calibration-controls}). Overall mixture dependence remains
because many pairs have different within-task performance orderings.
Normalization is not a new technique; LM-Polygraph itself studies quantile
normalization \citep{vashurin2025polygraph}. Our audit concerns fixed score
representations and explicit deployment mixtures, not an assertion that the
source benchmark published an erroneous pooled-AUROC leaderboard.

Exhaustive exact two-group stress testing detects \emph{every} raw primary
reversal on discovery, confirmation, and full banks. There are zero observed
pairs that pass all two-group continua but fail the full simplex. Thus the
proved support hierarchy is a theoretical and synthetic result, not an
observed advantage over exhaustive pairwise stress testing in this bank.

\subsection{Certified radii, descriptive uncertainty, and validation}

Table~\ref{tab:fixed-radii} reports exact-optimization intervals for the
minimum TV shift producing a 0.01 AUROC reversal. These are optimization
brackets, not statistical intervals. The near-minimal discovery witnesses do
not always transfer: pairs 1 versus 20 and 6 versus 23 have positive
confirmation gaps at those witnesses. The larger original stress witnesses
in Table~\ref{tab:fixed-examples} do transfer. For pair 22 versus 18, the
confirmation reference already favors $B$ by more than 0.01, giving radius
zero.

\begin{table}[ht]
\centering\small
\caption{One-point reversal radii for the five fixed examples. Last column
evaluates the discovery radius's feasible witness unchanged on confirmation.}
\label{tab:fixed-radii}
\begin{tabular}{lrrr}
\toprule
Pair & Discovery TV bracket & Confirmation TV bracket & Confirmation gap at discovery witness\\
\midrule
3 vs 31 & $[.290039,.291016]$ & $[.285156,.286133]$ & $-.012242$\\
1 vs 20 & $[.129883,.130859]$ & $[.166016,.166992]$ & $+.007247$\\
13 vs 10 & $[.115234,.116211]$ & $[.103516,.104492]$ & $-.013046$\\
6 vs 23 & $[.091797,.092773]$ & $[.141602,.142578]$ & $+.006090$\\
22 vs 18 & $[.083008,.083984]$ & $[0,0]$ & $-.035620$\\
\bottomrule
\end{tabular}
\end{table}

For descriptive uncertainty we resample input/story clusters independently
within each confirmation task, retain all rows in each sampled cluster, and
renormalize row mass within task. We use 2,000 replicates and seed 18172026.
The pointwise percentile intervals in Table~\ref{tab:fixed-examples} neither
control all 595 comparisons simultaneously nor establish that benchmark
clusters are iid deployment draws. Exact finite-bank statements do not
require that interpretation. Independent weighted ROC-AUC calculations using
scikit-learn reproduce the matrix calculations, including discrete ties and
zero-mass groups. Rank-preservation checks cover every strict calibration
variant, method, and task.

All pair tables, integer counts, exact certificates, protocols, parsers,
and figure code will be released with the code. Figures are computed from
observed scores, not simulated model outputs. A second-model, correctness-validated independent study remains
essential before generalizing these results to semantic reliability or
deployment populations. The present evidence establishes a reproducible
finite-bank audit and its calibration and endpoint boundaries.

All 595 primary-endpoint pairs and the first three fixed-pair mixture curves
are shown in Figures~\mref{fig:all-pair-audit}
and~\mref{fig:fixed-mixture-curves} of the main paper.

%% file: sec/supp_validation.tex
\section{Completed Synthetic and Comparator Validation}
\label{app:validation}
\subsection{Independent-Sampling Coverage and Power}
We use the exact hierarchy distribution in Theorem~\mref{thm:hierarchy},
with $K=3$, $\pi_g=1/2$, $c=17/24$, and scores
$S_A=4G+Y$, $S_B=0$ on negatives and $S_B\in\{-1,1\}$ on positives,
with positive probability $c$ for $+1$. Every repetition independently
generates $n$ observations per group. The true uniform-mixture gap is
$-1/24$. We compare the all-pairs Hoeffding bands at total error probability
$\delta=.05$ with their intersection with fixed-matching empirical-Bernstein
bands, using the split error allocation in Appendix~\ref{app:statistics}.
There is one prespecified score comparison, so $J=1$ and $L=6$.

The generator uses NumPy's default PCG64 generator with seed 1812026, and
500 independent repetitions at each of $n=500,2000,8000,32000$. The
configuration was fixed before the simulation was run. Coverage means that \emph{all six} true upper-triangular matrix entries lie
in the reported intervals. Power means that the upper matrix has a strictly
negative quadratic form at the prespecified uniform mixture; it is not
power optimized over mixtures. Both procedures cover all six entries in
all 500 repetitions at each sample size. The Wilson 95\% interval for each
coverage proportion is $[.9924,1]$. This limited simulation checks a
specific distribution and implementation; the coverage theorem, rather
than the simulation, supplies a distribution-free statement.

\begin{table}[ht]
\centering\small
\caption{Completed independent-sampling experiment. Counts are out of 500.
The half-width column is the median over repetitions of the largest matrix
entry interval half-width. Different interval centers are retained when
intersecting the two constructions.}
\label{tab:synthetic}
\begin{tabular}{rlrrrr}
\toprule
$n$ per group & Method & Joint coverage & Reversal certificates & 95\% interval for power & Half-width\\
\midrule
500 & Hoeffding & 500 & 0 & $[0,.0076]$ & .104696\\
500 & Intersection & 500 & 0 & $[0,.0076]$ & .106432\\
2000 & Hoeffding & 500 & 0 & $[0,.0076]$ & .052348\\
2000 & Intersection & 500 & 0 & $[0,.0076]$ & .037200\\
8000 & Hoeffding & 500 & 0 & $[0,.0076]$ & .026174\\
8000 & Intersection & 500 & 32 & $[.0457,.0890]$ & .014477\\
32000 & Hoeffding & 500 & 236 & $[.4286,.5158]$ & .013087\\
32000 & Intersection & 500 & 500 & $[.9924,1]$ & .006200\\
\bottomrule
\end{tabular}
\end{table}

At the smallest sample size, the intersection can be wider than the
standalone Hoeffding interval because its Hoeffding component receives only
half the error budget. At larger sizes the variance information tightens
the intervals. The substantial sample requirement is retained as a negative
finding: exact computation of the finite-bank mixture problem does not
remove the cost of distribution-free population inference.

\begin{figure}[ht]
\centering
\includegraphics[width=.93\textwidth]{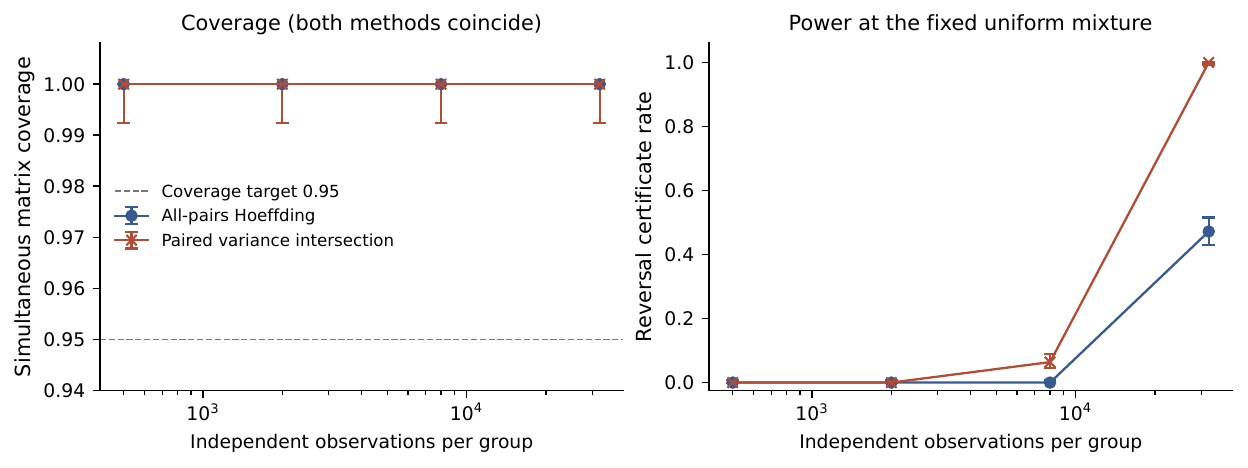}
\caption{Completed synthetic coverage and fixed-witness power. Points are
empirical proportions over 500 repetitions; intervals are Wilson 95\%
intervals. The target coverage is .95. All model and data in this figure are
the explicit synthetic construction; no point represents real-model
generalization.}
\label{fig:coverage-power}
\end{figure}

\subsection{Corrected Grid Against Exact Continuum Optimization}
We evaluate Theorem~\mref{thm:grid} using all denominator-40 grid points on
the full four-group simplex for every primary pair on discovery and
confirmation. Each grid value and lower certificate is computed exactly
from integer rank counts and rational prevalences. A nonnegative lower
bound certifies dominance; a negative feasible grid value certifies a
reversal; other cases remain unresolved. The exact continuum classification
uses the separate stationary-face optimizer. No grid minimum is called an
exact continuous normalized-gap minimum.

\begin{table}[ht]
\centering\small
\caption{Corrected-grid classifications versus exact classifications,
all 595 released-Accuracy pairs in each bank. ``Width'' is the interval
between the corrected lower bound and the feasible grid gap.}
\label{tab:grid-validation}
\setlength{\tabcolsep}{5pt}
\begin{tabular}{@{}lrrrrrr@{}}
\toprule
Bank & Grid dominance & Grid reversal & Unresolved & Exact dominance & Median width & Maximum width\\
\midrule
Discovery & 168 & 419 & 8 & 176 & .006837 & .018203\\
Confirmation & 156 & 422 & 17 & 173 & .006948 & .017595\\
\bottomrule
\end{tabular}
\end{table}

All 1,190 comparisons agree whenever the corrected grid reaches a decision;
there are no invalid decisions. Every negative continuum case has a grid
witness in this bank. This agreement is empirical, whereas the correction's
validity is proved for every admissible instance. The exact grid arithmetic
agrees with the separately computed floating grid values to within
$2.99\times10^{-16}$. All 1,190 grid runs together took 76.64\,s of
CPU time.

\subsection{Paired Hellinger and Total-Variation Comparators}
For each of the five discovery-selected illustrations, we evaluate the
paired-loss Hellinger bound in Eq.~\eqref{eq:hellingerbaseline} using exact
finite-bank moments at the confirmation reference mixture and the exact
group-pair affinity to the \emph{unchanged discovery} stress mixture.
We also evaluate the paired total-variation bound
$|\Delta(q)-\Delta(q_0)|\le2\TV(P_{q_0}^+\otimes P_{q_0}^-,
P_q^+\otimes P_q^-)$, with group identities retained. The factor two is
the range length of the paired gap loss. These are certificates for the
same paired comparison rather than separate, unnecessarily uncoupled
certificates for the two scores.

For the five full-simplex witnesses, both generic drift bounds include
zero and do not certify a comparison sign; the exact structured audit
finds the negative values in Table~\ref{tab:fixed-examples}. The five
interior-domain witnesses likewise remain unresolved by those bounds.
The complete numerical affinity, moments, lower and upper endpoints, and
observed gap for each of the ten cases will be released with the code.
These outcomes
illustrate the cost of relaxing known mixture structure to generic drift.
They do not claim a uniform superiority theorem, an optimal Hellinger-ball
implementation, or statistical population validity for empirical moments.

\begin{table}[ht]
\centering\small
\caption{Completed pointwise drift comparisons. Pair numbers follow the
five rows of Table~\ref{tab:fixed-examples}. The intervals use the exact
finite-bank paired-loss moments and group-pair drift, with outward-rounded
endpoints. All ten intervals include zero.}
\label{tab:metric-baselines}
\begin{tabular}{rlrrr}
\toprule
Pair & Discovery witness domain & Confirmation gap & Hellinger interval & TV interval\\
\midrule
1 & Full simplex & $-.1234$ & $[-1.000,1.000]$ & $[-1.000,1.000]$\\
2 & Full simplex & $-.0793$ & $[-1.000,1.000]$ & $[-1.000,1.000]$\\
3 & Full simplex & $-.1017$ & $[-.991,.994]$ & $[-1.000,1.000]$\\
4 & Full simplex & $-.0609$ & $[-.997,1.000]$ & $[-1.000,1.000]$\\
5 & Full simplex & $-.1148$ & $[-.999,.998]$ & $[-1.000,1.000]$\\
1 & $q_g\ge.05$ & $-.0584$ & $[-.850,.964]$ & $[-.971,1.000]$\\
2 & $q_g\ge.05$ & $-.0475$ & $[-.770,.850]$ & $[-1.000,1.000]$\\
3 & $q_g\ge.05$ & $-.0737$ & $[-.722,.741]$ & $[-1.000,1.000]$\\
4 & $q_g\ge.05$ & $-.0406$ & $[-.743,.786]$ & $[-1.000,1.000]$\\
5 & $q_g\ge.05$ & $-.0904$ & $[-.776,.760]$ & $[-1.000,1.000]$\\
\bottomrule
\end{tabular}
\end{table}

\subsection{Finite-Bank Alignment Envelope Checks}
For every pair, we compute ordinary two-sample KS distances exactly from
the empirical score distributions and use them conservatively in
Theorem~\mref{thm:alignment}. Ordinary KS upper-bounds mid-CDF distance:
the right-limit CDF difference and its left limit both lie in the same
supremum interval, hence so does their average. All matrix endpoints are
rational. Across 595 pairs, 16 ordered entries per matrix, and five
representations (raw, marginal mid-CDF, marginal strict rank,
negative-class mid-CDF, negative-class strict rank), all 47,600 matrix
entries satisfy the deterministic envelope. The numbers of universal
$A$-dominance certificates are respectively $0,82,83,0,0$.
The last two zeros concern the marginal-discrepancy envelope, whose
assumption is not targeted by negative-class calibration. They are not
an assessment of the best possible negative-class-alignment certificate.

The raw-to-strict-marginal drop in the median method's maximum pairwise
group KS distance is $.78561685$ to $.0843629$. A small discrepancy can
make the envelope informative without exact equality of group marginals.
All these are deterministic empirical-law comparisons, separate from the
unlabeled calibration-sample guarantee in Appendix~\ref{app:alignment}.

%% file: sec/supp_protocols.tex
\section{Essential Independent Validation and Optional Extension}
\label{app:protocols}
E1 has not been run; its protocol is stated below and is not evidence for
any claim in the paper. Its purpose is to determine whether the practical
conclusions extend beyond the released base-model bank and its literal-match
labels. The vision study O1 (Appendix~\ref{app:o1results}) and an auxiliary
smaller-model language replication E2 (Appendix~\ref{app:e2results}), which
does not replace E1, were run.

\subsection{E1 (Essential, Not Run): Independent Model, Validated Outcomes, and Decision Consequences}
\label{app:essential}
\paragraph{Questions.}
The prespecified questions are: (i) do population-mixture comparisons remain
useful for an independently generated model under task-native correctness;
(ii) how much of any observed instability survives marginal alignment and
exhaustive two-group stress tests; (iii) do certificates and reversal radii
predict held-out comparison behavior; and (iv) does the chosen comparison
change a global error-review decision at a fixed review budget? No particular
answer is assumed. A second model is replication of the audit question, not
a paired comparison of two models' distinct error labels.

\paragraph{Pinned model and data.}
Use Qwen3-8B with its shipped tokenizer and chat template at a pinned
checkpoint revision \citep{qwen3model}, thinking mode disabled, bfloat16
weights, and no fine-tuning or quantization. The following Hugging Face
dataset snapshots are used at pinned revisions.
\begin{center}\small
\begin{tabular}{ll}
\toprule
Task & Hugging Face dataset/configuration\\
\midrule
MMLU & \texttt{cais/mmlu}, per-subject configurations\\
GSM8K & \texttt{openai/gsm8k}, \texttt{main}\\
CoQA & \texttt{stanfordnlp/coqa}, default\\
TriviaQA & \texttt{mandarjoshi/trivia\_qa}, \texttt{rc.nocontext}\\
\bottomrule
\end{tabular}
\end{center}
These tasks are MMLU \citep{hendrycks2021mmlu}, GSM8K
\citep{cobbe2021gsm8k}, CoQA \citep{reddy2019coqa}, and TriviaQA
\citep{joshi2017triviaqa}. The exact revisions and software environment
are recorded with the code to be released.

\paragraph{Deterministic selection and independence of decisions.}
Assign stable row identifiers using the task name and the dataset's provided
ID, or the SHA-256 of its complete input and target JSON when no ID exists.
Define a global cluster identifier as the SHA-256 of the complete input text,
using the shared story for CoQA. Normalize CRLF and CR to LF, strip leading
and trailing whitespace, and encode as UTF-8; do not otherwise normalize or
add a task prefix. Thus identical inputs across tasks and all questions from
one CoQA story remain in the same cluster. For subset selection sort clusters
lexicographically by the digest of
\texttt{p181-selection-v1} + NUL + task name + NUL + global cluster ID,
encoded as UTF-8. Break any hash tie lexicographically by cluster ID.
For evaluation use the first 100 MMLU test questions per subject (5,700
total if every subject has sufficient rows), all 1,319 GSM8K test problems,
the first 2,000 distinct TriviaQA validation questions, and every question
in the first 250 CoQA validation stories. If a source has fewer eligible
units than requested, retain all and report the shortfall; do not replace
it based on scores or labels. Keep original question multiplicity inside
selected CoQA stories. Reserve independent calibration data: up to 10
MMLU validation questions per subject, the first 1,000 GSM8K training
questions, 1,000 TriviaQA training questions, and the first 100 CoQA
training stories under the same hash rule. Exclude any exact input overlap
between calibration and evaluation, report it, and do not refill it.

Use an independently domain-separated split digest: SHA-256 of the UTF-8 encoding of
\texttt{p181-replication-v1} + NUL + global cluster ID. Values below 128
in its first byte give discovery, otherwise confirmation. Do not include a
task name in this split digest or reuse the subset-selection digest.
Calibration data fit transforms
only; discovery selects pair orientation and witness mixtures; confirmation
evaluates the unchanged selections. Preserve every scored row in the
finite-bank estimand. Training-data contamination of public benchmarks by
the pretrained model is not excluded by this protocol and must be listed as
a limitation; it does not create a claim of new training generalization.

\paragraph{Prompts and generation.}
Use a single user message and no task-specific system prompt. The task
messages are exactly the following templates, with bracketed fields replaced
by dataset text:
\begin{quote}\small
MMLU: \texttt{[question]\\{A. [choice A]}\\{B. [choice B]}\\{C. [choice C]}\\{D. [choice D]}\\{Answer with only A, B, C, or D.}}\par\smallskip
GSM8K: \texttt{[question]\\{Solve the problem. End with "The answer is [number]".}}\par\smallskip
TriviaQA: \texttt{[question]\\{Answer with a short phrase only.}}\par\smallskip
CoQA: \texttt{Story: [story]\\{[history]Question: [current question]}\\{Answer with a short phrase only.}}
\end{quote}
Every displayed line break inside a template is one LF character. CoQA
\texttt{[history]} is empty at the first turn; otherwise concatenate, in
dataset order, \texttt{Question: [previous question]} + LF +
\texttt{Answer: [previous reference answer]} + LF for each preceding turn,
using the dataset's primary reference answer. No additional blank lines or
spaces are inserted. The MMLU prefix is the original question followed by
the four original choices in order; the symbolic \texttt{[number]} in the
GSM8K instruction is literal text, not replaced before generation.
The greedy answer determines the binary target for every uncertainty method.
Use deterministic greedy generation, no sampling, minimum new tokens 2,
maximum new tokens 16
(MMLU), 512 (GSM8K), and 64 (CoQA and TriviaQA); stop on EOS. Save token IDs,
token log probabilities, entropy ingredients, decoded answer, stop reason,
and truncation flag. Truncated or unparsable answers remain in the study and
are scored by the declared task rule. For sampling-based uncertainty methods,
draw 10 additional answers at temperature 1.0, top-$p=1.0$, and top-$k=50$,
seed 181, with the same token limits. Sampling answers do not replace the greedy
target. Hash and retain all outputs before computing new comparison tables.

\paragraph{Score methods and fairness of comparison.}
The essential panel is MaximumSequenceProbability, Perplexity,
MeanTokenEntropy, CCP, PTrue, SemanticEntropy, and SAR: all 21 unordered
pairs. Use the LM-Polygraph implementations at a pinned source revision,
with the same default estimator definitions. The two auxiliary checkpoints
are \texttt{microsoft/deberta-large-mnli} (CCP and SemanticEntropy) and
\texttt{cross-encoder/stsb-roberta-large} (SAR), at pinned revisions.
PTrue uses the
same Qwen checkpoint and the source's default True-token prompt, without
an external judge. Use sequence CCP with 10 token alternatives;
SemanticEntropy with \texttt{class\_probability\_estimation=sum},
\texttt{entropy\_estimation=mean}, and
\texttt{use\_unique\_responses=False}; and SAR with $t=0.001$.
Do not substitute an estimator if it fails. The raw-score panel, marginal-rank control, and negative-class
rank control use identical greedy outputs and labels. Marginal transforms
use calibration scores without labels; negative-class transforms use only
calibration examples declared correct by the task rule. Both controls use
the strict piecewise-linear mid-rank interpolation with positive-slope end
extrapolation defined in Appendix~\ref{app:empirical-details}, fitted only
on that independent calibration sample. The implementation uses exact
rational interpolation
to preserve close-score ordering. If integer scores are needed for rank
counting, encode the order of all exact transformed values with one common
map per method across groups; this preserves every comparison and does not
change the fitted pointwise function. No confirmation-dependent tie repair
or refitting is allowed. No score orientation
is chosen from confirmation AUROC. If a method fails on any row, report the
failure counts and cause, retain the planned panel as incomplete, and show
both a common-complete-case sensitivity and the unchanged remaining methods;
do not silently delete a failing method from the registered family.

This panel includes token-level, verbalized, semantic-sampling, and
relevance-weighted scores. The contribution is the comparison audit, so
winning against these uncertainty estimators is not a success criterion.
The audit baselines are pooled AUROC at $q_0$, within-task and macro AUROC,
entrywise cross-AUC dominance, all two-group continua, uncorrected and
corrected grids with $m\in\{10,20,40\}$, exact continuum optimization,
the marginal-alignment envelope, and the paired Hellinger/TV comparator.
All use the same allowed mixtures and score pairs. If a task-by-split bank
contains only one class, mark its within-task AUROC and the full-simplex
audit under the paper's positivity assumption as undefined. Retain every
row; do not resample, reassign splits, or relabel examples to create both
classes. The implementation rejects such inputs. Separately reported
restricted-domain analyses must state their changed estimand and admissible
mixture region. If a calibration task has no correct examples, mark its
negative-class rank control unavailable rather than borrowing labels from
discovery or confirmation.

\paragraph{Primary correctness and human audit.}
Use first valid leading answer-letter extraction for MMLU, exact final
numeric-answer comparison for GSM8K, official normalized alias exact match
for TriviaQA, and official multi-reference CoQA token F1 thresholded at
$0.5$. The continuous CoQA F1 and thresholds $0.25,0.75,1.0$ are
prespecified sensitivities. Parser failure is incorrect. For numerical
answers, remove thousands separators and compare exact Decimal values;
use the final numeric token after the final \texttt{The answer is} marker,
or the final numeric token if the marker is absent. No judge model defines
the primary label.

The pinned Hugging Face CoQA export omits additional reference answers and
original conversation IDs. For validation scoring only, join it to
the official CoQA development set using exact story text,
ordered question texts, and ordered primary-answer texts. A story-only
join is insufficient. The completed source-consistency check uniquely
matches all 500 validation conversations and 7,983 turns; each turn has
four reference entries. The official evaluation script computes turn F1 by averaging the maximum over references after leaving
each reference out in turn. Additional references enter neither prompts
nor subset selection. Training calibration uses its available single
reference and the evaluator's single-reference branch. The join report will be released with the code. This check
validates reference availability, not model answers.

Freeze all eligible confirmation row identifiers and operational labels
before annotation. Within each task-by-label stratum sort the stable row
identifiers lexicographically. Using one PCG64 generator with seed
18172027, iterate tasks in the order MMLU, CoQA, GSM8K, TriviaQA, and then
labels $0,1$; in a stratum of size $N$, draw $\min(100,N)$ indices uniformly
without replacement, giving at most 200 answers per task. Save the sampled
identifiers before annotation; do not replace difficult cases. Randomize the
pooled presentation order with a separate generator (seed 18172028) and
omit stratum and operational-label fields.
Two independent annotators see the question, required context, references,
and answer, but no uncertainty scores, mixture, method names, or operational
label. They label task correctness as correct, incorrect, or indeterminate;
a third adjudicator resolves disagreements. Retain indeterminate cases and
report optimistic and pessimistic sensitivity bounds rather than deleting
them. Report agreement, adjudication counts, and label error rates weighted
back to stratum sizes. For a finite-bank interval, invert the hypergeometric
sampling tails within each of the eight task-by-label strata, allocating
$.05/16$ to each one-sided tail, then combine the resulting stratum bounds with
known stratum weights. Do not treat balanced annotation sampling as a
representative unweighted population.

If the audit finds more than 5\% weighted mislabeling in a task, or shows a
systematic answer-format failure, adjudicate that task's entire confirmation
bank before making a semantic-error claim. Freeze the repaired label rule
without inspecting uncertainty-comparison outcomes and report the original
and adjudicated results side by side. If full adjudication is not completed,
restrict the conclusion to the operational task metric. No favorable
comparison under an inaccurate label is counted as validation of
human-error detection.

\paragraph{Mixture regions, witnesses, and primary endpoints.}
Use $q_0=(1/4,1/4,1/4,1/4)$. Report the full simplex, $q_g\ge0.05$, and
their intersections with TV radii $r\in\{0.05,0.10,0.20,0.30\}$ about
$q_0$. The full simplex is a stress domain, not an asserted deployment law.
The bounded-share TV regions are declared deployment scenarios; no claim
that a particular organization experiences them is permitted without data.
For all 21 pairs report exact finite-bank dominance, explicit reversals,
worst-gap brackets of width at most $10^{-4}$, and reversal-radius brackets
of width at most $10^{-3}$ for $\kappa\in\{0.01,0.02\}$. Select pair
orientation from the raw discovery AUROC at $q_0$ and retain that orientation
in all three score panels; break exact ties by the listed method order.
For each panel and declared region, select a discovery near-minimizer of
the normalized AUROC gap with a certified optimality bracket of width at
most $10^{-4}$. On confirmation evaluate these exact same mixtures before
any separate reoptimization. Include unchanged and
reoptimized results as different columns. A one-point negative gap is the
prespecified materiality threshold, not a significance threshold.

\paragraph{Review-budget consequence.}
At budgets $b\in\{0.05,0.10,0.20\}$, route the largest-score fraction $b$
to review under each mixture, using random fractional allocation at the
threshold to attain exactly the budget. On a finite bank, row $i$ in group
$g$ has mass $q_g/n_g$. Compute error recall among reviewed mass, error
rate among accepted mass, and change from the equal-task mixture. This is
a transparent simulation of a fixed review budget, not a deployed human
study. Evaluate raw and calibrated scores at $q_0$ and at the unchanged
discovery AUROC near-minimizer selected for each panel and declared region.
A TV radius specifies a region, not an evaluation mixture. These are
AUROC-selected mixtures; no review-budget objective is optimized over a
region or used to select a confirmation mixture. AUROC reversal does not logically
imply reversal of any one budget metric; agreement and disagreement must
both be reported. Do not tune the review budget to produce agreement.

\paragraph{Uncertainty and resources.}
Report pointwise paired cluster bootstrap intervals (2,000 resamples, seed
181) for discovery-fixed confirmation witnesses and review-budget contrasts,
resampling stories or duplicate-input clusters within tasks. These are
approximate conditional intervals. Do not bootstrap score pairs as
independent observations. For formal population statements, separately
define an iid cluster sampling target, take one prespecified hash-selected
question per independent cluster, and apply the simultaneous envelope at
total $\delta=.05$ across all 63 comparisons (21 pairs in each of the raw,
marginal-rank, and negative-class-rank panels). The simultaneous mixture
envelope already covers every declared region and adaptive witness within
that family. CoQA label-threshold sensitivities are descriptive unless
additional multiplicity allocation is explicitly made before evaluation.
This changes the target from
row-weighted to cluster-representative performance and must be reported as
such. If independence is not defensible, report no population certificate.
Do not lower the family size or enlarge the error budget after seeing
inconclusive intervals. Report the number of exact solves and all
unresolved optimizer stops. Run generation once; repeated greedy seeds are
not independent model replications.

\paragraph{How outcomes change the claims.}
If the independent study shows material comparison changes inside the
bounded TV scenarios and unchanged-witness transfer under validated labels,
the paper may claim practical mixture sensitivity beyond the released bank.
If marginal alignment again removes all hidden within-task-win reversals,
retain the tractable-boundary interpretation; this supports the mechanism
and does not establish a universally useful raw-score failure. If every
reversal is found by two-group checks, keep the higher-support hierarchy as
theory only. If raw reversals disappear under correctness validation, make
no semantic-error claim. If population envelopes remain mostly unresolved,
state the sample-efficiency limitation and keep population guarantees as
conditional results. If AUROC changes do not affect the fixed review
budgets, do not claim downstream utility. A null second-model result limits
empirical scope; it does not falsify the theorems, and no post-hoc tasks are
added to obtain a favorable replication.

\subsection{O1: A Second Domain and More Groups}
\label{app:o1protocol}
Use frozen torchvision ResNet-50 \texttt{IMAGENET1K\_V2} weights on the
first 1,000 ImageNet validation image IDs under the domain-separated
\texttt{p181-selection-v1} subset-selection hash,
and ImageNet-C severities $1,3,5$ \citep{hendrycks2019imagenetc}.
The 15 standard corruption types are
Gaussian noise, shot noise, impulse noise, defocus blur, glass blur, motion
blur, zoom blur, snow, frost, fog, brightness, contrast, elastic transform,
pixelate, and JPEG compression. Define $Y=1$ when top-1 classification is
wrong. Compare negative maximum softmax probability, softmax entropy,
negative top-two logit margin, and negative log-sum-exp of logits at fixed
temperature one, giving six pairs. Apply the shipped weight preprocessing
and save logits; do not refit the classifier. Partition selected image IDs
using SHA-256 of the UTF-8 encoding of \texttt{p181-vision-split-v1} + NUL
+ image ID. Interpret the first eight digest bytes as an unsigned big-endian
integer and take its remainder modulo three for calibration, discovery,
and confirmation, respectively. These partitions have approximately equal
sizes; report their actual sizes. All corrupted versions of one image
remain together, and the split digest is independent of subset selection.

At each fixed severity treat the corruption types as groups; separately
report the prespecified noise, blur, weather, and digital family aggregation
($K=4$) and the full $K=15$ analysis. Apply exact stationary-face and
branch-and-bound audits to the four-family setting. For $K=15$, report only
the corrected grids $m=2,3,4$: lower and upper brackets, explicit witnesses,
grid-point counts, CPU times, and unresolved cases. The exact polytope
construction is intended for small $K$ and is not applied at $K=15$. The same image-cluster kernel in
Appendix~\ref{app:statistics} governs any population interval. Report raw
and marginal-rank controls, all pair counts, certificate widths, finite
bank reversals, processing times, and coverage only under an explicit
independent-image sampling model. This evaluates domain transfer and
scaling. Do not infer robustness to unseen corruption distributions.

\paragraph{O1 result.}
\label{app:o1results}
These results are also reported in Sec.~\mref{sec:vision} and
Table~\mref{tab:o1} of the main paper; the complete record follows.
Inference over all 45{,}000 corrupted-image evaluations (verified official
archives) was run once. For $K=15$, the general vertex oracle would enumerate
$\binom{30}{14}$ bases of the unconstrained simplex, which is impractical, so
the aggregation uses the closed form for the unconstrained simplex (its
vertices are the unit vectors), which agrees exactly with the oracle on small
cases. This substitution was made after inference; no protocol setting,
threshold, or split changed. The split produced 342 calibration, 316 discovery, and
342 confirmation images. Each of the 18 comparisons (six score pairs at three
severities) is oriented once by raw discovery AUROC and kept fixed in both
panels; a reversal is material at 0.01 AUROC.

\input{generated/o1.tex}
Wins counts all-group winners on discovery; Cert.\ counts confirmation
continuum certificates; Rev.\ witness counts explicit confirmation reversal
witnesses; Unres.\ counts unresolved comparisons. Frozen rev.\ counts,
among discovery winners, reversals of the frozen discovery witness on
confirmation, with material
ones in parentheses; for $K=15$ only the material frozen-grid count is
computed (--).

For the four-family setting ($K=4$), 15 of 18 comparisons are all-family
winners on discovery. All 15 are certified as continuum winners on
confirmation, the three non-winners carry explicit confirmation reversal
witnesses, and no frozen discovery witness reverses. Under image-cluster
resampling (2{,}000 replicates; pointwise, descriptive, not a population
certificate) every raw-panel winner keeps its 95\% interval above $-0.01$ at
the frozen witness; in the marginal-rank panel 14 of 15 do. For all 15
corruptions ($K=15$), 14 of 18 comparisons win every group on discovery; the
14 winners give 10 confirmation certificates, 3 unresolved comparisons and one
explicit confirmation reversal (severity 5, negative top-two margin against
negative log-sum-exp; confirmation gap at most $-0.050$ for $m=2,3,4$); the four
non-winners give three reversal witnesses and one unresolved comparison. No
frozen discovery witness reverses materially. Raw and
marginal-rank controls give identical counts. Each audit used under 10\,s of
CPU. This is one frozen model on one corruption benchmark; it says nothing
about unseen corruption distributions.

\subsection{E2: Auxiliary Smaller-Model Language Replication (Not E1)}
\label{app:e2results}
E2 runs the E1 code path (prompts, generation budgets, the seven score
methods, all three panels, mixture regions, frozen raw-discovery
orientation, and the 0.01 materiality threshold) with three deviations.
The model is the smaller Qwen3-1.7B in float16 (pinned revision) instead of
Qwen3-8B in bfloat16. Selection is capped at 250 MMLU,
GSM8K and TriviaQA questions and 60 CoQA stories for evaluation (100, 100,
100 and 30 for calibration), giving 1{,}715 evaluation rows instead of E1's
12{,}948. Discovery and confirmation use a separate split salt, so E2's
splits do not coincide with E1's. E2 was added after the E1 protocol was
frozen, uses operational correctness labels only (no human audit), and
cannot satisfy or replace any E1 endpoint. All 2{,}526 rows completed: 811
for calibration and 1{,}715 for evaluation (863 discovery, 852 confirmation),
with 21 method pairs per panel. All computations used one pinned
configuration; model generations and NLI scores for all 2{,}526 rows and SAR
scores for 1{,}999 rows were computed before an interruption, and the
remaining 527 SAR rows and every analysis step after resuming. No row was
generated twice.

\input{generated/e2.tex}

Error rate is the share of evaluation rows labelled incorrect. Wins counts
pairs that win within every task on discovery. Frozen rev.\ counts, among
those winners, confirmation reversals at the frozen full-simplex discovery
witness (material ones, at most $-0.01$, in parentheses). Univ.\ cert.\
counts the pairs (of 21) for which the method ranked first on raw discovery
at $q_0$ is certified on the confirmation bank to be at least as good at
every mixture. This is a finite-bank certificate; no population certificate
is computed. In every panel these are the three P(True) comparisons and
semantic entropy over CCP, which does not win within every task. In both
aligned panels three further pairs are certified in the reverse direction.
Median radius is the median discovery upper bound on $r_{0.01}$
(Eq.~\meqref{eq:radius}) over the pairs where it is finite (16, 18 and 16 of
21; the rest have $r_{0.01}=+\infty$). A value of 0 means the discovery gap at
$q_0$ is already at or below $-0.01$ under the raw-discovery orientation. In
each aligned panel 9 of 21 pairs are at 0, so the medians are not comparable
across panels. Boot.\ CI counts winners whose pointwise 95\% interval at the
frozen witness lies entirely above or entirely below zero. The intervals come
from resampling input or story clusters within each task on the confirmation
split (2{,}000 replicates); they are descriptive, not simultaneous.

Four comparisons win every task on discovery: maximum sequence probability
over CCP, and perplexity, mean token entropy, and SAR each over P(True). The
three P(True) comparisons are finite-bank certified on the confirmation
split at every mixture in all three panels. Three caveats apply. No
population certificate is computed. Under marginal alignment all three are
mixture dependent on discovery (minima $-0.032$, $-0.039$, $-0.038$). At
the frozen witness, the pointwise bootstrap interval excludes zero for 1, 0
and 1 of the three in the raw, marginal-rank and negative-class-rank panels.

Maximum sequence probability over CCP is mixture dependent on confirmation.
Its frozen discovery witness puts all weight on GSM8K and TriviaQA (about
0.52 and 0.48) and none on MMLU or CoQA. At that witness the confirmation
gap is $-0.037$ for raw scores, $-0.026$ after marginal strict-rank
alignment, and $-0.002$ after negative-class strict-rank alignment. The
reversal is therefore material in the first two panels and not in the
third. Under cluster resampling the raw interval is $[-0.058,-0.014]$,
entirely below $-0.01$. The marginal-rank interval $[-0.056,0.005]$ and the
negative-class-rank interval $[-0.027,0.022]$ include zero. So, on a second
model, a within-task winner again loses at a mixture chosen on discovery and
held fixed. On the completed Stable LM 2 12B bank, marginal alignment removed all
66 raw within-task-winner reversals (of 239 winners). Here it does not remove
the single raw reversal as a point estimate ($-0.026$), although that
interval includes zero.

Three limits bound this result. First, the frozen MMLU budget of 16 new
tokens is too short for this model, which usually does not begin with an
option letter, the only form the parser accepts. 238 of 255 MMLU
evaluation rows hit the budget, and 169 could not be parsed (169 of the 193
MMLU errors), so MMLU labels here mostly measure answer-format compliance.
The headline reversal's witness gives MMLU no weight. Second, TriviaQA uses
exact alias matching and labels 81.2\% of rows incorrect. Third, labels are
operational, with no human audit. E2 shows that the reversal phenomenon
appears on a second model, smaller than the E1 model and from the same
family, on a capped and differently split selection. It is not evidence
about Qwen3-8B, and E1 remains essential.

\paragraph{Implementation status.}
The audit, grid correction, marginal and metric envelopes, calibration
controls, statistical intervals, and generic score-table interface are
implemented and tested. The E1 model generation and human annotations were
not completed; no output from them is used. O1 and E2 were run and are
reported above. The replication configuration will be released with the
code; no result is filled by an expectation or simulated model output.

%% file: generated/e2.tex
\begin{center}\small
\begin{tabular}{lrrrr}
\toprule
Task & Eval.\ rows & Error rate & Parser fail. & Trunc.\\
\midrule
MMLU & 255 & 75.7\% & 169 & 238\\
CoQA & 960 & 44.2\% & 0 & 4\\
GSM8K & 250 & 17.6\% & 0 & 6\\
TriviaQA & 250 & 81.2\% & 0 & 0\\
\bottomrule
\end{tabular}
\par\medskip
\begin{tabular}{lrrrrr}
\toprule
Panel & Wins & \shortstack{Frozen rev.\\(material)} & Univ.\ cert. & \shortstack{Median\\radius} & \shortstack{Boot.\ CI\\$>0$ / $<0$}\\
\midrule
Raw & 4/21 & 1 (1) & 4 & 0.219 & 1 / 1\\
Marginal rank & 4/21 & 1 (1) & 4 & 0.027 & 0 / 0\\
Neg.-class rank & 4/21 & 1 (0) & 4 & 0.000 & 1 / 0\\
\bottomrule
\end{tabular}
\end{center}

%% file: sec/supp_limitations_appendix.tex
\section{Scope and Limitations}
\label{app:limitations}
The realizability reduction and the support hierarchy are mathematical
existence results; the grid, optimization, statistical, and alignment bounds
hold under their stated assumptions; and the empirical results are exact
properties of finite released score banks. We do not claim a new uncertainty
estimator, AUC decomposition, copositivity algorithm, or a first certificate
for shifted AUC. The full simplex is a stress domain rather than a realistic
deployment law, exact methods become expensive for many groups,
distribution-free population intervals can be conservative at benchmark
sizes, and marginal rank alignment can remove genuine differences in base
risk and is not a fairness criterion. The independent-model study E1
(Appendix~\ref{app:essential}) has not been run.